\documentclass[journal]{IEEEtran}
\usepackage{blindtext}
\usepackage{graphicx}
\usepackage{amssymb}
\usepackage{graphicx}
\usepackage{bbm}
\usepackage{mathtools}
\usepackage{mathrsfs}

\usepackage{booktabs}
\usepackage{multirow}
\usepackage{algorithm}
\usepackage{algorithmic}
\usepackage{bbding}
\usepackage[table,xcdraw]{xcolor}
\usepackage{epsfig}
\usepackage{xcolor}
\usepackage{pifont}
\usepackage{amsmath}
\usepackage{subfigure}

\usepackage{hyperref}
\newcommand{\citet}[1]{\citeauthor{#1} \shortcite{#1}}

\newlength\savewidth
\newcommand\shline{\noalign{\global\savewidth\arrayrulewidth
                            \global\arrayrulewidth 0.8pt}%
                   \hline
                   \noalign{\global\arrayrulewidth\savewidth}}

\makeatletter
\DeclareRobustCommand\onedot{\futurelet\@let@token\@onedot}
\def\@onedot{\ifx\@let@token.\else.\null\fi\xspace}

\def\eg{\emph{e.g}\onedot} 
\def\ie{\emph{i.e}\onedot}

\makeatother

\ifCLASSINFOpdf
\else
\fi
\begin{document}
%

\title{\textbf{COAST}: \textbf{CO}ntrollable \textbf{A}rbitrary-\textbf{S}ampling Ne\textbf{T}work for Compressive Sensing}

\author{Di You,~Jian Zhang,~Jingfen Xie,~Bin Chen,~Siwei Ma 

\thanks{D. You, J. Zhang, J. Xie, B. Chen are with the School of Electronic and Computer Engineering, Peking University Shenzhen Graduate School, Shenzhen 518055, China. J. Zhang is also with the Peng Cheng Laboratory, Shenzhen 518052, China. (e-mail: diyou@pku.edu.cn; zhangjian.sz@pku.edu.cn; xiejingfenn@163.com; chenbin74851@163.com)

S. Ma is with the School of Electronics Engineering and Computer Science, Peking University, Beijing, 100871, China. (e-mail: swma@pku.edu.cn)
}
}

\markboth{2021}%
{Shell \MakeLowercase{\textit{et al.}}: Bare Demo of IEEEtran.cls for Journals}

\maketitle

\begin{abstract}
\noindent 
Recent deep network-based compressive sensing (CS) methods have achieved great success. However, most of them regard different sampling matrices as different independent tasks and need to train a specific model for each target sampling matrix. Such practices give rise to inefficiency in computing and suffer from poor generalization ability. In this paper, we propose a novel COntrollable Arbitrary-Sampling neTwork, dubbed COAST, to solve CS problems of arbitrary-sampling matrices (including unseen sampling matrices) with one single model. Under the optimization-inspired deep unfolding framework, our COAST exhibits good interpretability. In COAST, a random projection augmentation (RPA) strategy is proposed to promote the training diversity in the sampling space to enable arbitrary sampling, and a controllable proximal mapping module (CPMM) and a plug-and-play deblocking (PnP-D) strategy are further developed to dynamically modulate the network features and effectively eliminate the blocking artifacts, respectively. Extensive experiments on widely used benchmark datasets demonstrate that our proposed COAST is not only able to handle arbitrary sampling matrices with one single model but also to achieve state-of-the-art performance with fast speed. The source code is available on \href{https://github.com/jianzhangcs/COAST}{https://github.com/jianzhangcs/COAST}.
\vspace{-12pt}
\end{abstract}

\section{Introduction}

As a typical inverse problem, compressive sensing (CS) aims to recover an unknown natural signal from a small number of its measurements acquired by a linear random projection, which has been successfully applied in single-pixel imaging \cite{duarte2008single, rousset2017adaptive}, accelerating magnetic resonance imaging (MRI) \cite{lustig2007sparse},  sparse-view computed tomography (CT) \cite{szczykutowicz2010dual}, wireless telemonitoring \cite{zhang2013compressed} and cognitive radio communication \cite{sharma2016application}.

Mathematically, supposing that ${\mathbf x} \in \mathbb{R}^N$ is the original natural signal and $\mathbf{\Phi}\in \mathbb{R}^{M \times N}$ is a linear random projection (sampling matrix), the CS measurement of ${\mathbf x}$, denoted by ${\mathbf y}\in \mathbb{R}^M$, is usually formulated as 
\begin{equation}
    \mathbf{y = \Phi (x + n)},
\label{eq: CS}
\end{equation}
where $\mathbf{n}$ denotes the additive  white Gaussian noise with standard  deviation $\sigma$. By default, $\sigma=0$ and $\mathbf{y}$ is called `noiseless'. The purpose of CS is to infer ${\mathbf x}$ from its randomized CS measurement ${\mathbf y}$. Because $M \ll N$, this inverse problem is typically ill-posed, whereby CS ratio, denoted by $\gamma$, is defined as $\gamma=\frac{M}{N}$. This paper mainly focuses on CS reconstruction of natural images, where  ${\mathbf x} \in \mathbb{R}^N$ is a vectorized representation of an image patch of size $\sqrt{N}$$\times$$\sqrt{N}$. It is worth noting that our proposed framework can be easily extended to various types of data, \textit{e.g.}, MRI and video data \cite{yang2016deep, zhao2016video}.

\begin{figure}[t]
\setlength{\abovecaptionskip}{0.cm}
\setlength{\belowcaptionskip}{0cm}
\centering
\includegraphics[width=0.95\linewidth]{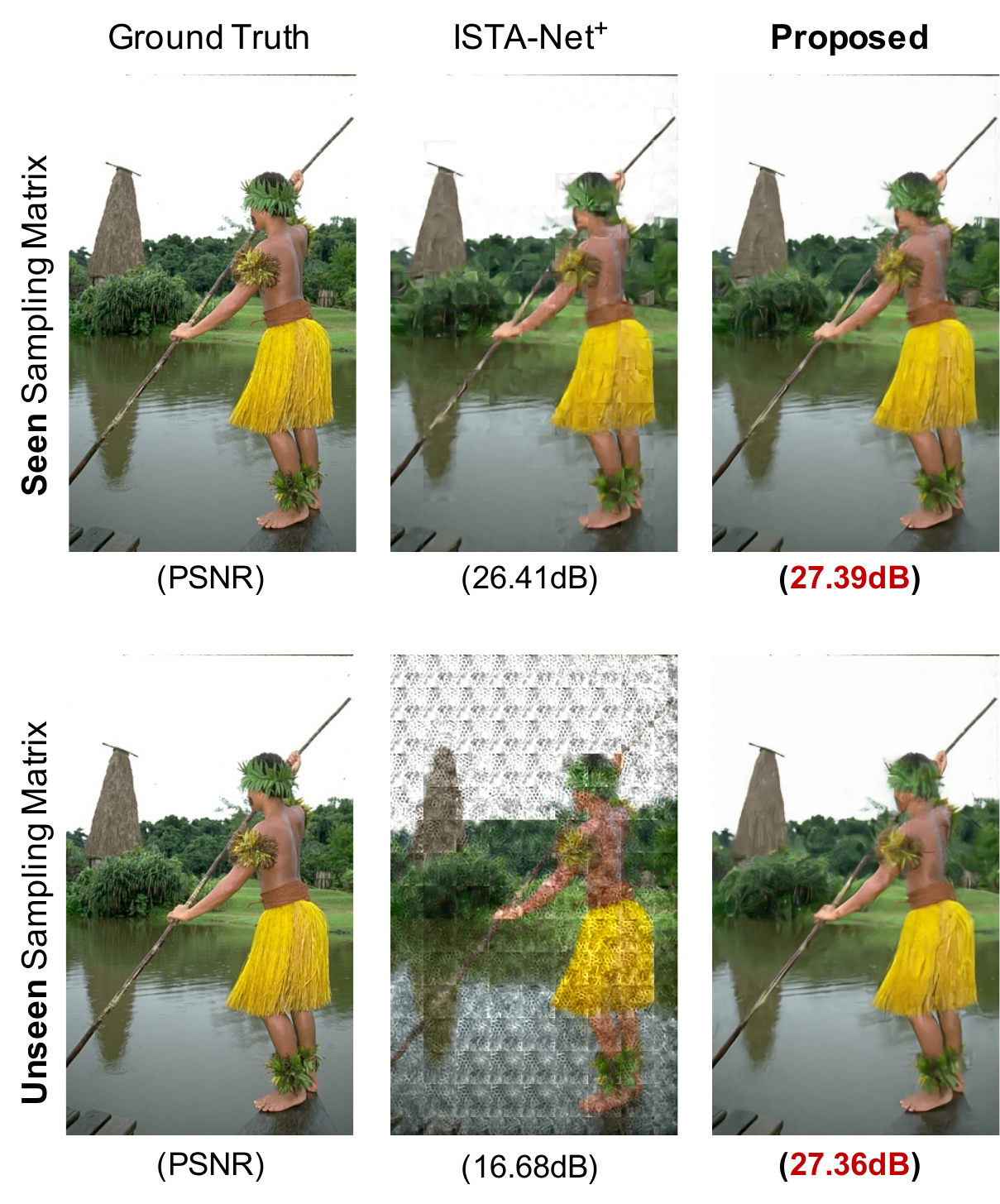} 
\caption{CS reconstruction results from the seen and unseen sampling matrices ($\gamma = 10\% $) produced by a recent state-of-the-art method, \textit{i.e}., ISTA-Net$^{+}$ \cite{zhang2018ista}, and our proposed COAST. It is clear to observe that COAST significantly surpasses the ISTA-Net$^{+}$, especially for the unseen sampling matrix.}
\vspace{-14pt}
\label{fig:introduction_example}
\end{figure}

In the past decade, a great number of image CS reconstruction methods, which include traditional model-driven methods and data-driven deep neural methods, have been developed. The traditional model-driven methods \cite{duarte2009learning, hong2018online} commonly convert the reconstruction problem to a Maximum a Posteriori (MAP) estimation problem and adopt an iterative manner to search for the optimal solution that can fit both the given measurement and prior constraints. Although these model-driven methods enjoy the advantage of interpretability, due to the iterative nature of the solutions and the hand-crafted characteristics, they suffer from high computational complexity and are difficult to characterize various complicated image structures. Fueled by the rise of deep learning, several data-driven deep neural methods \cite{adler2016deep, du2018fully, shi2017deep,shi2019image, lohit2018convolutional,shi2019scalable,9199540,chen2020learning} have been recently proposed for image CS reconstruction by direct learning the inverse mapping from the CS measurement domain to the original signal domain. Compared to the model-driven methods, the data-driven deep neural methods dramatically reduce time complexity due to their non-iterative nature of the solutions and achieve impressive reconstruction performance. However, they are often trained as a black box, with limited insights from the CS domain. Most recently, to address the above drawbacks of both the model- and data-driven methods, some optimization-inspired explicable deep networks are proposed \cite{zhang2018ista, gregor2010learning, 8753511, DBLP:conf/nips/MetzlerMB17, borgerding2017amp, yang2016deep, dong2019denoising}, which combine the merits of both the model- and data-driven methods, yielding a better signal recovery performance.

\begin{figure}[t]
\setlength{\abovecaptionskip}{0.cm}
\setlength{\belowcaptionskip}{0cm}
\centering
\includegraphics[width=0.9\linewidth]{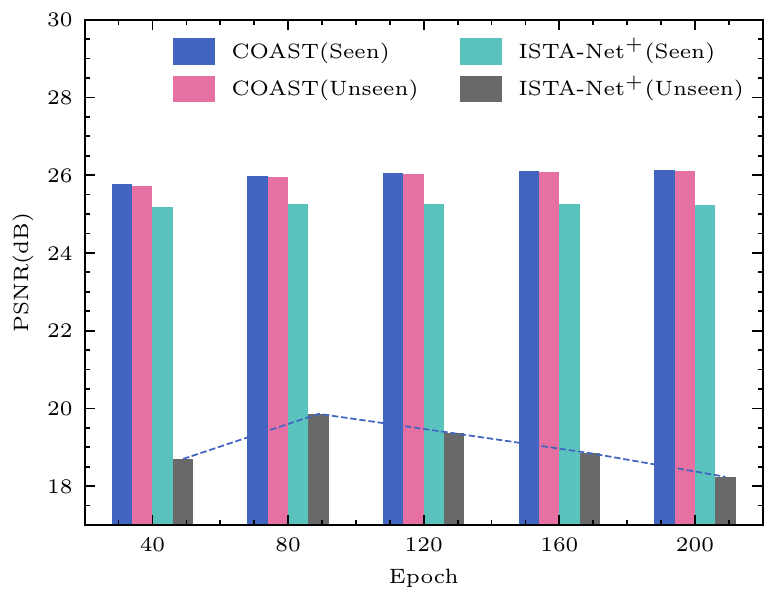} 
\caption{PSNR values of ISTA-Net$^{+}$ and our COAST on BSD68 dataset with respect to various training epochs in the case of $\gamma = 10\% $. It can be observed that for the seen sampling matrix, both ISTA-Net$^{+}$ and COAST can be improved by the training process. For the unseen sampling matrix, the reconstruction performance of ISTA-Net$^{+}$ decreases with the increase of training epochs, while our COAST still achieves very similar performance with the seen sampling matrix.}
\vspace{-10pt}
\label{fig:introduction_overfit}
\end{figure}

Although existing deep network-based methods have achieved impressive results, most of them regard different sampling matrices as different independent tasks and need to train a specific model for each target sampling matrix with fixed CS ratio and patch size.
For the convenience of discussion, we consider these methods as \textit{$\mathbf{\Phi}$-specific} methods, due to the fact that they need to be re-trained for various sampling matrices.
In Fig.~\ref{fig:introduction_example}, we give an example and make a comparison between a state-of-the-art CS model, \textit{i.e.}, ISTA-Net$^{+}$ \cite{zhang2018ista} and our COAST. As we can see, ISTA-Net$^+$ can produce a desirable result for the specific target (seen) sampling matrix, but can not handle other (unseen) sampling matrix. As shown in the green and gray bars of Fig.~\ref{fig:introduction_overfit}, we can see that ISTA-Net$^+$ tend to overfit to the specific data used for training. Specifically, with the increasing of the epoch number, its PSNR performance increases on the seen sampling matrix but decreases on the unseen sampling matrix.
In brief, we can learn that current deep network-based models trained on a specific sampling matrix suffer from poor generalization ability and can not handle arbitrary sampling matrices well with a single model. For different sampling matrices, they usually need to re-train different network models separately. Obviously, such practices result in large storage space and high time complexity, and  do not satisfy the needs of the real scenarios that usually contain various sampling matrices.

In this paper, to address the above issues, we propose a novel \textbf{CO}ntrollable \textbf{A}rbitrary-\textbf{S}ampling ne\textbf{T}work, dubbed \textbf{COAST}, for compressive sensing. As a \textit{$\mathbf{\Phi}$-agnostic} method, our COAST is only trained once and can handle arbitrary sampling matrices (including unseen sampling matrices) with one single model. In particular, COAST consists of a sampling subnet (SS), an initialization subnet (IS), and a recovery subnet (RS). A random projection augmentation (RPA) strategy in SS is proposed as the core component to handle arbitrary sampling problem with one single model. Moreover, a controllable proximal mapping module (CPMM) and a plug-and-play deblocking (PnP-D) strategy are proposed to improve the robustness and enhance the performance of COAST. 
Extensive experiments fully verify the effectiveness of our COAST. Specifically, for the arbitrary seen sampling matrices, our COAST is able to achieve competitive results with the corresponding single models, which are re-trained for different sampling matrices.
For the arbitrary unseen sampling matrices, COAST can achieve very similar performance with the corresponding seen sampling matrices (see the Gray bar in Fig.~\ref{fig:introduction_overfit}).
More encouragingly, as shown in Fig.~\ref{fig:introduction_overfit}, COAST significantly outperforms the previous state-of-the-art  ISTA-Net$^{+}$ \cite{zhang2018ista}, especially for the unseen sampling matrices.

Overall, the main contributions of this paper are as follows:
\begin{itemize}
\item  We propose a novel \textbf{CO}ntrollable \textbf{A}rbitrary-\textbf{S}ampling ne\textbf{T}work (\textbf{COAST}) for CS, which is able to handle arbitrary sampling with one single model.

\item A random projection augmentation (RPA) strategy is developed to promote the generalization ability and the performance of networks, which can also be directly incorporated into existing deep network-based CS methods.

\item A controllable proximal mapping module (CPMM) and a plug-and-play deblocking (PnP-D) strategy are further proposed to further improve the robustness and enhance the reconstruction quality.

\item Our COAST achieves state-of-the-art performance for arbitrary sampling matrices with one single model, promoting the application in real-world CS systems.

\end{itemize}

\section{Background}

This section will present some recent related works, and briefly introduce traditional iterative shrinkage-thresholding algorithm (ISTA) \cite{blumensath2009iterative} for facilitating the follow-up discussion.

\subsection{Related works}
We generally group existing compressive sensing (CS) reconstruction methods of images into two categories: traditional optimization-based CS methods and recent network-based CS methods. In what follows, we give a brief review of both and focus on some recent work most relevant to our own.

\textbf{Optimization-based Methods:}
Given the input linear measurements $\mathbf{y = \Phi (x + n)}$, traditional image CS methods usually reconstruct the original image $\mathbf{\hat{x}}$ by solving the following optimization problem:
\begin{equation}
\underset{{\mathbf{\hat{x}}}}{\min} \frac{1}{2}\|\mathbf{\Phi \mathbf{\hat{x}}- \mathbf{y}}\|^2_2 + \lambda{\psi({\mathbf{\hat{x}})}},
\label{eq: classical CS}
\end{equation}
where $\mathbf{\hat{x}}$ and $\mathbf{\Phi}$ denote the recovered image patch and linear random projection (sampling matrix), respectively, and $\psi(\mathbf{\hat{x}})$ denotes the hand-crafted image prior-regularized term with $\lambda$ being the (generally pre-defined) regularization parameter.

Optimization-based Methods are based on regularization terms inspired by image priors. In the early stage, the prior knowledge about transform coefficients (\eg statistical dependencies \cite{kim2010compressed,he2009exploiting}, sparsity \cite{zhang2014imageSP}, etc.) are exploited to reconstruct visual images. Furthermore, some elaborate priors exploiting the non-local self-similarity properties of natural images have been proposed to improve CS reconstruction \cite{zhang2013improved,DBLP:journals/tip/ZhangZG14,DBLP:journals/esticas/ZhangZZXMG12,DBLP:conf/dcc/ZhaoZM016,dong2014compressive}. Recently, some fast and effective convolutional neural network (CNN) denoisers are trained and integrated into Half Quadratic Splitting (HQS) \cite{zhang2017learning} and alternating direction method of multipliers (ADMM) \cite{chang2017projector, zhao2018cream} to solve image inverse problems.
However, all these traditional image CS methods require hundreds of iterations by means of various iterative solvers (\eg ISTA \cite{blumensath2009iterative}, ADMM \cite{li2013efficient}, or AMP \cite{metzler2016denoising}), which inevitably increase computational cost thus restricting the application of CS. In addition, the selected image prior (\eg optimal transform) or the optimization parameters (\eg step size and regularization parameter) are usually hand-crafted and quite challenging to pre-define.

\textbf{Network-based Methods:} Recently, inspired by the powerful learning capability of deep networks and its success in computer vision tasks, some network-based methods have been proposed. \cite{DBLP:conf/allerton/MousaviPB15} first proposes to apply a stacked denoising auto-encoder (SDA) to learn the representation from training data and to reconstruct test data from the CS measurements. \cite{adler2016deep} and \cite{DBLP:journals/dsp/IliadisSK18} separately propose to use fully-connected neural networks for image and video CS reconstruction. \cite{kulkarni2016reconnet} further develops a CNN-based CS method, dubbed ReconNet, which learns to regress an image patch (output) from its CS measurement (input). In order to erase the blocking artifacts and further enhance the performance, \cite{shi2017deep} proposes a sampling-recovery framework called CSNet. Moreover, a joint sampling and scalable reconstruction CS network is proposed, dubbed SCSNet, allowing the one well-trained model to deal with several carefully-designed sampling matrices \cite{shi2019scalable}. However, SCSNet can not be generalized to arbitrary sampling matrices.

Lately, some deep unfolding networks, \eg, ADMM-Net \cite{yang2016deep}, ISTA-Net \cite{zhang2018ista}, DPDNN \cite{DBLP:journals/pami/DongWYSWL19}, and OPINE-Net \cite{9019857}, are proposed to combine the merits of both optimization-based and network-based methods. Specifically, \cite{yang2016deep} proposes a so-called ADMM-Net architecture by reformulating ADMM for CS magnetic resonance imaging (CS-MRI) using deep networks. ISTA-Net \cite{zhang2018ista}, which works well for both general CS and CS-MRI based on ISTA, goes beyond that to adopt nonlinear transforms to more effectively sparsify natural images and develops an efficient strategy for solving their proximal mapping problems. DPDNN \cite{DBLP:journals/pami/DongWYSWL19} is designed to solve common inverse problems by unfolding the denoising process inspired by HQS method. Another deep unfolding model dubbed OPINE-Net \cite{9019857} maps an optimization problem into the deep network for joint adaptive binary sampling and recovery of image CS for the first time.
In addition, based on the idea that the structure of the auto-encoder network itself is a good prior to capture image statistics, an unsupervised method DIP \cite{ulyanov2018deep} is proposed to directly fit the network for each corrupted image. To further improve the performance of DIP and overcome its dependence on the number of iterations, NLR-CSNet \cite{8999514} includes non-local prior in addition to deep network prior and constructs an HQS-based optimization method for network learning.

\textbf{Weakness:}
As we can see, recent research on CS has achieved great success due to the rapid development of deep convolutional neural networks. However, the CS problem of arbitrary sampling matrices has been ignored for a long time. Most existing works regard different sampling matrices as different independent tasks. To handle arbitrary sampling matrices, a) exiting traditional optimization-based methods usually need to adopt an iterative manner to re-search for the optimal solution that can fit each specific sampling matrix; and b) most of network-based methods need to re-train a network model separately for each sampling matrix, otherwise it will suffer from poor performance on the new given sampling matrix (see Fig.~\ref{fig:introduction_example}). Such re-search and re-train practices inevitably result in both time-consuming and storage-consuming issues.

\begin{figure*}[t]
\centering
\includegraphics[width=1.0\textwidth]{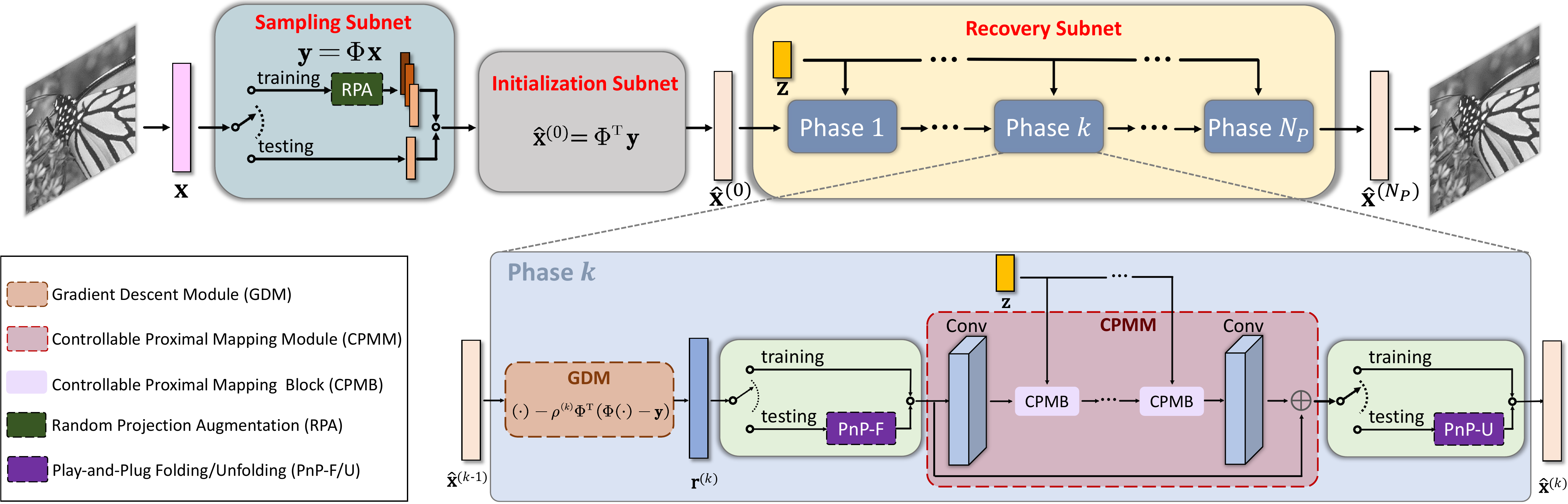} 
\caption{Illustrations of training and testing processes for our proposed COAST network, which consists of a sampling subnet (SS), an initialization subnet (IS), and a recovery subnet (RS).
}
\label{fig: opticsframework}
\vspace{-6pt}
\end{figure*}

\subsection{Traditional ISTA for CS}

The iterative shrinkage-thresholding algorithm (ISTA) is a popular first-order proximal method, which is well suited for solving many large-scale linear inverse problems. Specifically, ISTA solves the CS reconstruction problem in Eq.~\eqref{eq: classical CS} by iterating between the following update steps:
\begin{align} \label{eq: r}
\mathbf{r}^{(k)} &= \mathbf{\hat{x}}^{(k-1)} - \rho \mathbf{\Phi}^{\top} (\mathbf{\Phi} \mathbf{\hat{x}}^{(k-1)} - \mathbf{{y}}), \\
\label{eq: x}
\mathbf{\hat{x}}^{(k)} &=  \underset{\mathbf{\hat{x}}}{\arg\min} ~\frac{1}{2}\|\mathbf{\hat{x}} - \mathbf{r}^{(k)}\|^2_2  + \lambda{\psi({\mathbf{\hat{x}})}},
\end{align}
where $k$ is the ISTA iteration index and $\rho$ is the step size. Eq.~\eqref{eq: r} is trivial, while Eq.~\eqref{eq: x} is actually a special case of the so-called proximal mapping, \textit{i.e.}, $\mathbf{prox}_{\lambda\psi}(\mathbf{r}^{(k)})$. Mathematically, the proximal mapping of regularizer $\psi$ denoted by $\mathbf{prox}_{\lambda\psi}(\mathbf{r})$ is defined as:
\begin{equation} \label{eq: x2}
\mathbf{prox}_{\lambda\psi}(\mathbf{r}) =  \underset{\mathbf{\hat{x}}}{\arg\min} ~\frac{1}{2}\|\mathbf{\hat{x}} - \mathbf{r}\|^2_2 +\lambda\psi(\mathbf{\hat{x}}).
\end{equation}

How to handle $\mathbf{prox}_{\lambda\psi}(\mathbf{r})$ in an efficient and effective way is of great importance for ISTA.

\section{Approach}
In this section, we elaborate on the design of our proposed COAST. We first formulate our COAST in Section~\ref{sec:formulate}. Then, as illustrated in Fig.~\ref{fig: opticsframework}, our COAST consists of a sampling subnet (SS), an initialization subnet (IS) and a recovery subnet (RS), which are described in Section~\ref{sec:arch}. Finally, we describe the training details in Section~\ref{sec:loss}.

\subsection{Problem Formulation}
\label{sec:formulate}
In CS of natural images, the common practice is to partition an image into several non-overlapped image patches and sample each patch independently with the same sampling matrix. Assuming we have an original image patch ${\mathbf x}$ and a sampling matrix $\mathbf{\Phi}$, we get the CS measurement $\mathbf{y}=\mathbf{\Phi}(\mathbf{x}+\mathbf{n})$, where $\mathbf{n}$ denotes the additive white Gaussian noise with standard deviation $\sigma$.
In order to improve reconstruction performance and increase network capacity, instead of hand-crafted, we make the image prior-regularized term $\psi(\cdot)$ as learnable parameters, and adopt a residual neural network to solve its proximal mapping operator effectively. Therefore, we obtain the following general problem formulation:
\begin{equation}
\underset{\mathbf{\hat{x}},  \psi}{\min} ~\frac{1}{2}\|\mathbf{\Phi} \mathbf{\hat{x}} - \mathbf{y}\|^2_2 + \lambda  \psi(\mathbf{\hat{x}}) ,
\label{eq:formulation}
\end{equation}
where $\mathbf{\hat{x}}$ denotes the recovered image patch.

\subsection{Architecture Design of COAST} \label{sec:arch}
To  map the optimization in Eq.~\eqref{eq:formulation} into deep network efficiently, we propose to implement it in a three-step scheme. Concretely, the first step is to simulate the sampling process and promote the training diversity. Then, the second step is to deal with the dimensionality mismatch between the original image and its CS measurement. The final step is to design the recovery network to generate the reconstructed image.

Accordingly, as illustrated in Fig.~\ref{fig: opticsframework}, our COAST consists of three subnets: a sampling subnet (SS), an initialization subnet (IS) and a recovery subnet (RS).

\subsubsection{\textbf{Sampling Subnet (SS)}} \label{sec:SS}
As illustrated in Fig.~\ref{fig: opticsframework}, we incorporate the proposed random projection augmentation (RPA) strategy into the sampling subnet (SS) to boost the generalization ability and the reconstruction performance. Here, we describe the SS at testing and training stages, respectively.

\textbf{Testing Stage:} RPA is not used at testing stage of SS. Similar to previous works \cite{kulkarni2016reconnet, zhang2018ista}, in SS we just employ the sampling matrix $\mathbf{\Phi} \in \mathbb{R}^{M \times N}$ to acquire the randomized CS measurement $\mathbf{y} = \mathbf{\Phi}(\mathbf{x+n}) \in \mathbb{R}^M$ from $\mathbf{x} \in \mathbb{R}^N$, which is a vectorized representation of a $\sqrt{N} \times \sqrt{N}$ image patch. 

\textbf{Training Stage:}
Similar with the popular data augmentation techniques \cite{DBLP:conf/iccv/YunHCOYC19}, our RPA on sampling matrices is only adopted at the training stage of SS. In the following, we will describe the training stage of SS and detail the RPA on how and why it works. 

\begin{figure}[t]
\centering
 \includegraphics[width=1\linewidth]{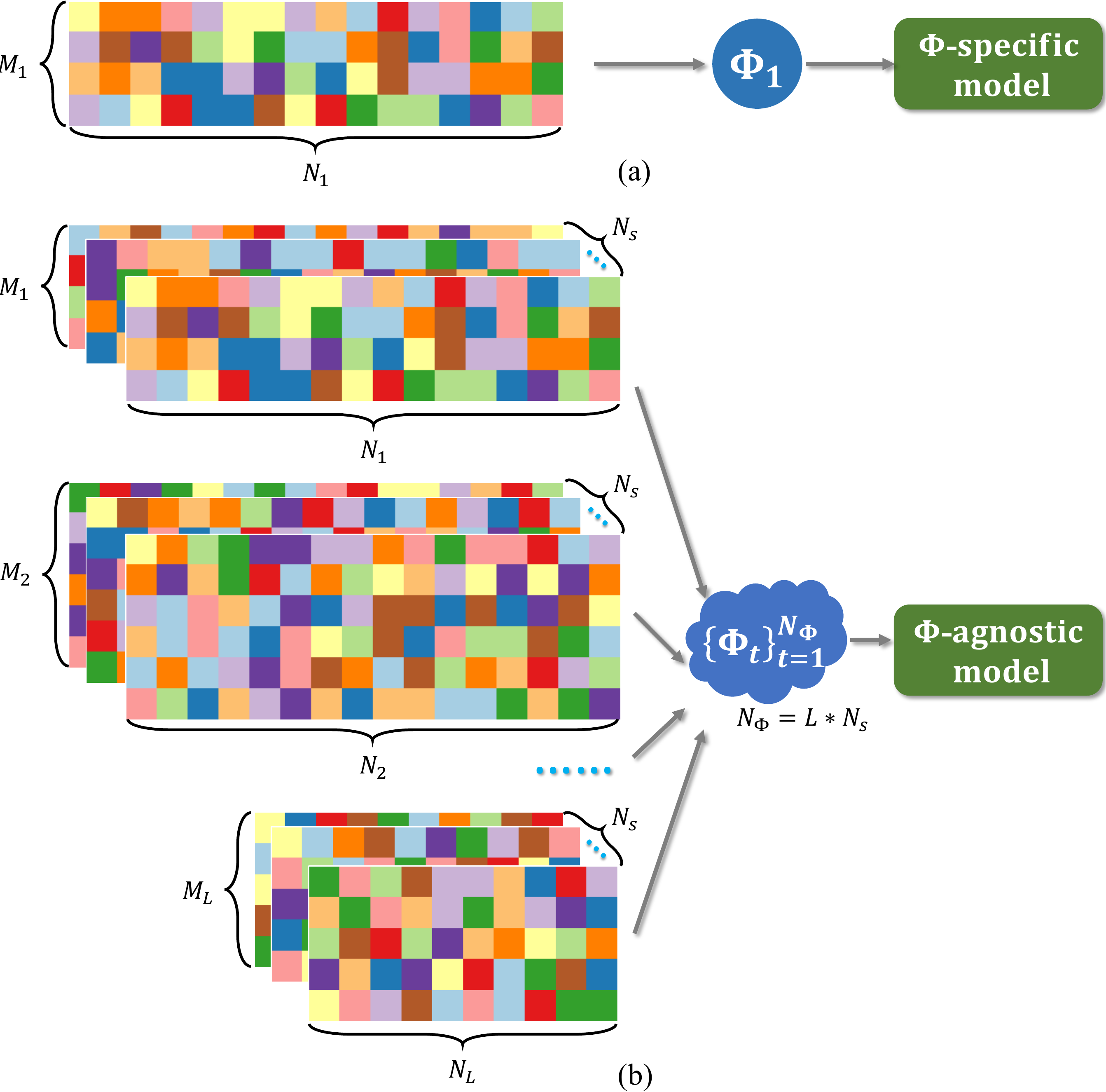} 
\caption{Comparison of $\mathbf{\Phi}$-specific and $\mathbf{\Phi}$-agnostic models. (a) previous approaches trained for specific sampling matrix ($\mathbf{\Phi}$-specific models) and (b) our approach trained for arbitrary sampling matrix ($\mathbf{\Phi}$-agnostic model).
The $M_t,N_t$ represent the dimensions of each sampling matrix.
As we can see, the previous approaches only accept one sampling matrix to learn $\mathbf{\Phi}$-specific models, while our RPA augments the sampling matrices by introducing additional sampling matrices to enable our COAST to handle arbitrary sampling matrices with one single model.}
\label{fig: RPA}
\vspace{-4pt}
\end{figure}

In previous works, as shown in Fig.~\ref{fig: RPA}(a), $\mathbf{\Phi}_1\in \mathbb{R}^{M_1 \times N_1}$ is a fixed sampling matrix used for training and testing, which means that the resulting network model is $\mathbf{\Phi}$-specific.

For our COAST, as shown in Fig.~\ref{fig: RPA}(b), the proposed RPA augments the training by introducing additional different sampling matrices. For a given sampling matrix with dimension $M$-by-$N$, denoted by $\mathbf{\Phi}_{M\times N}$, we introduce the $\mathcal{RPA}$ operator by adopting similar ways to generate its augmented set, including $N_S$ sampling matrices with the same dimension as $\mathbf{\Phi}_{M\times N}$. Thus, the proposed $\mathcal{RPA}$ operator is defined as: 
\begin{equation}
\mathcal{RPA}\left(\mathbf{\Phi}_{M\times N} \right) =\left\{\mathbf{\Phi} _{M\times N\circledast 1},\cdots ,\mathbf{\Phi}_{M\times N\circledast N_S} \right\},
\end{equation}
where $\mathbf{\Phi} _{M\times N\circledast 1}=\mathbf{\Phi} _{M\times N}$ and $\mathbf{\Phi} _{M\times N\circledast k} \neq \mathbf{\Phi}_{M\times N}, 2\leq k \leq N_S$.
Furthermore, the $\mathcal{RPA}$ operator is extended to deal with a set of multiple sampling matrices with $L (L\geq 2)$ different arbitrary dimensions, \textit{i.e.}, arbitrary CS ratios and patch sizes, which is defined as follows:
\begin{equation}
\begin{aligned}
& \mathcal{RPA}\left( \left[ \mathbf{\Phi}_{M_1\times N_1}, \dots, \mathbf{\Phi}_{M_L \times N_L} \right] \right) = \\
& \mathcal{RPA}\left(\mathbf{\Phi}_{M_1\times N_1} \right)  \cup \dots \cup \mathcal{RPA}\left(\mathbf{\Phi}_{M_{L}\times N_{L}} \right) .
\end{aligned}
\end{equation}
Accordingly, the sampling matrices augmented by RPA form a set $\{\mathbf{\Phi}_t\}_{t=1}^{N_{\Phi}}$, where the total number of the augmented sampling matrices is $N_\Phi = L \times N_S$. It is worth emphasizing that our RPA can be applied to two widely-used types of sampling matrices, \textit{i.e.} \textit{fixed random Gaussian matrix (FRGM)} and \textit{data-driven adaptively learned matrix (DALM)}, which are respectively elaborated as follows:


\begin{itemize}
\item 
For \textbf{FRGM}, each sampling matrix $\mathbf{\Phi}$ is constructed by generating a random Gaussian matrix and then orthogonalizing its rows, \ie $\mathbf{\Phi}\mathbf{\Phi}^{\top} = \mathbf{I}$, where $\mathbf{I}$ is the identity matrix. In this case, we implement our RPA by\textit{} generating multiple FRGMs using different random seeds multiple times for argumentation. 
\item 
For \textbf{DALM}, each sampling matrix $\mathbf{\Phi}$ is regarded as a learnable network parameter and is adaptively learned by jointly optimizing the sampling matrix and the reconstruction operator with training dataset. In practical implementation, we implement the proposed RPA by utilizing OPINE-Net \cite{9019857} multiple times to generate multiple DALMs for argumentation.
\end{itemize}

Now, we introduce how to incorporate the proposed RPA into our CS model. With RPA, we extend the fixed sampling matrix $\mathbf{\Phi}_1$ (see Fig.~\ref{fig: RPA}(a)) to its augmented set $\{\mathbf{\Phi}_t\}_{t=1}^{N_\Phi}$ (see Fig.~\ref{fig: RPA}(b)) and acquire the CS measurement $\mathbf{y}_{t}=\mathbf{\Phi}_{t}(\mathbf{x}+ \mathbf{n})$. So we can obtain the following optimization framework with the proposed RPA by replacing the fixed $\mathbf{\Phi}$ in Eq.~(\ref{eq:formulation}) with $\{\mathbf{\Phi}_t\}_{t=1}^{N_\Phi}$, which is formulated as:
\begin{equation}
\underset{\mathbf{\hat{x}},  \psi}{\min}\sum_t^{N_\Phi} \left\{\frac{1}{2}\|\mathbf{\Phi}_t \mathbf{\hat{x}} -\mathbf{y}_{t}\|^2_2 + \lambda  \psi(\mathbf{\hat{x}})\right\}.
\label{eq:analysis_model}
\end{equation}
In the practical implementation of casting Eq.~\eqref{eq:analysis_model} into deep network, we randomly select a sampling matrix from $\{\mathbf{\Phi}_t\}_{t=1}^{N_{\Phi}}$ for each batch. Accordingly, in the following, the subscript $t$ is omitted without confusion for simplicity. 

Compared with Eq.~\eqref{eq:formulation}, Eq.~\eqref{eq:analysis_model} makes the CS model $\mathbf{\Phi}$-agnostic. On one hand, RPA augments the training and alleviates the over-fitting problem.
On the other hand, the learning from multiple augmented data by RPA will significantly improve the \textit{generalization} capability of network. Besides, RPA can enable our COAST to jointly learn different knowledge from different sampling matrices which further promotes the \textit{performance} of each sampling matrix. In brief, the RPA enjoys two advantages: promoting the \textbf{generalization} capability of CS models and promoting the \textbf{performance} of each sampling matrix. The effectiveness of RPA will be validated in the experimental section.

\subsubsection{\textbf{Initialization Subnet (IS)}} \label{sec:IS}

Similar to traditional ISTA, our COAST also requires an initialization, which is denoted by $\mathbf{\hat{x}}^{(0)}$ in Fig.~\ref{fig: opticsframework}. In particular, to conduct the initialization process, previous network-based methods, such as \cite{adler2016deep, kulkarni2016reconnet}, usually exploit the parameterized fully-connected layer, due to the dimensionality mismatch between the original image and its CS measurement. However, such practice inevitably gives rise to a high amount of parameters and restricts the CS models to only accept the input measurements generated by the specific sampling matrix with the specific \textit{fixed dimension}.
In order to enable our CS model to handle arbitrary sampling matrices without introducing new parameters, in this paper, we propose to initially set zero values to $\mathbf{\hat{x}}^{(0)}$ and set $\mathbf{\rho}^{(0)}$ to be 1, \textit{i.e.} $\mathbf{\hat{x}}^{(0)} = 0$. Interestingly, this is equivalent to set $\mathbf{\hat{x}}^{(0)} = \mathbf{\Phi}^{\top}\mathbf{y}$. The reason is that both the initializations of $\mathbf{\hat{x}}^{(0)}$ and $\mathbf{\rho}^{(0)}$ will lead to the same $\mathbf{r}^{(1)}$, \textit{i.e.} $\mathbf{r}^{(1)} = \mathbf{\Phi}^{\top}\mathbf{y}$, according to Eq.~(\ref{eq: 
r2}). Intuitively, the design of our IS is a naive solution to handle various sampling matrices with \textit{arbitrary dimension}.

\begin{figure}[t]
\centering
\includegraphics[width=0.9\linewidth]{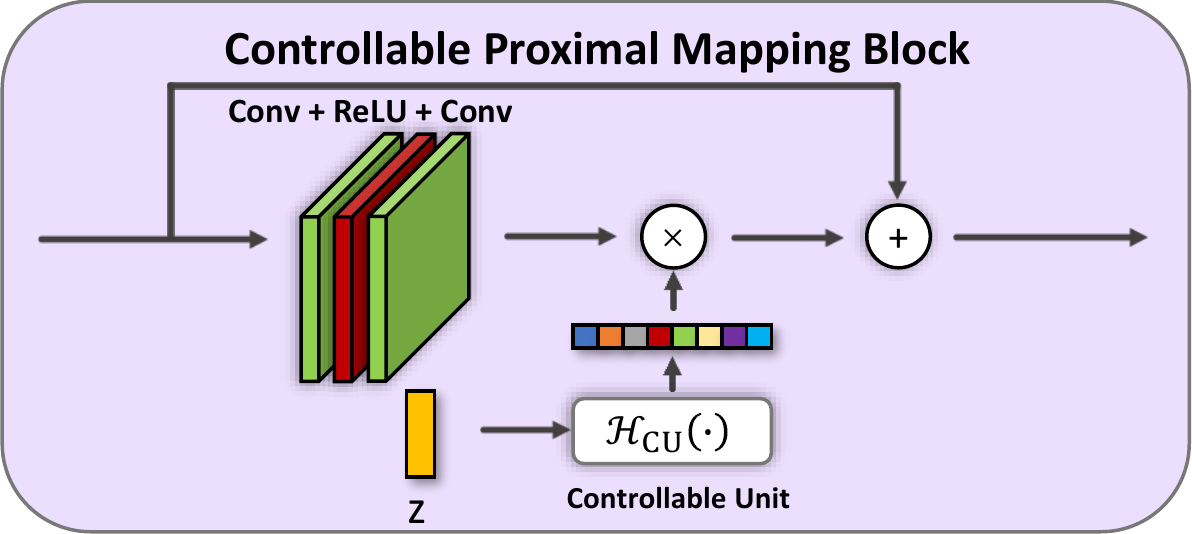}
\caption{Illustration of the proposed controllable proximal mapping block (CPMB), which introduces a controllable unit (CU) into a standard residual block. Specifically, the CU, which implemented by a fully-connected layer, takes the condition vector $\mathbf{z}$ as input to generate the controllable vector, which is used to dynamically modulate the output features of the residual block.}
\vspace{-4pt}
\label{fig:PMB}
\end{figure}

\subsubsection{\textbf{Recovery Subnet (RS)}}
\label{sec:recovery}

Considering the simplicity and interpretability, we design a deep unfolding model named Recovery Subnet (RS) to solve  Eq.~\eqref{eq:formulation}. In this paper, we map traditional ISTA into deep network to construct our RS as \cite{zhang2018ista}. It is worth noting that as a deep unfolding model, the RS structurally enables our COAST to handle arbitrary sampling matrices with one single model. Specifically, compared with classical deep network methods, deep unfolding methods enable the sampling matrix to play a more important role, which gives more potential of boosting the generalization ability on arbitrary sampling matrix.

As illustrated in Fig.~\ref{fig: opticsframework}, the recovery subnet consists of $N_P$ phases and each phase is composed of a  gradient descent module (GDM) and a controllable proximal mapping module (CPMM), which are corresponding to the two update steps in traditional ISTA, \textit{i.e.} Eq.~(\ref{eq: r}) and Eq.~(\ref{eq: x}). Besides, at the front and back ends of CPMM, we incorporate the proposed plug-and-play deblocking (PnP-D) strategy at the testing stage.

\begin{figure*}[t]
\centering
\includegraphics[width=1.0\textwidth]{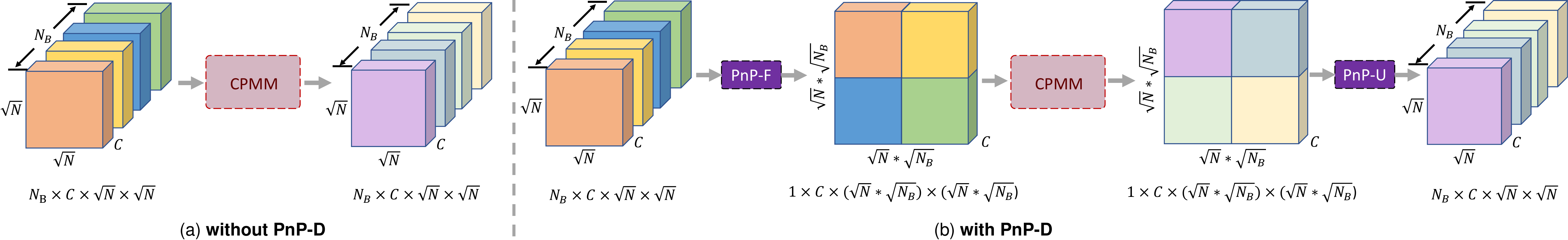}
\caption{Comparison of (a) the model `without PnP-D' and (b) our model `with PnP-D' at the testing stage. The $N_B$ and $\sqrt{N}$ represent the block number of each image and the size of the image patch, respectively. The $C$ denotes the channel size of the feature maps. 
}
\label{fig: ppd}
\vspace{0pt}
\end{figure*}

\textbf{Gradient Descent Module (GDM):}
To preserve the ISTA structure, gradient descent module (GDM) is directly defined according to Eq.~\eqref{eq: r}, where the step size $\rho$ becomes a learnable parameter. Taking $\mathbf{\hat{x}}^{(k-1)},\mathbf{\Phi},\mathbf{{y}}$ as input, the output of GDM in $k$-th phase, denoted by $\mathcal{H}_\text{GDM}^{(k)}$, is finally defined as:
\begin{equation}
\begin{aligned} \label{eq: r2}
&\mathbf{r}^{(k)}  = \mathcal{H}_\text{GDM}^{(k)}(\mathbf{\hat{x}}^{(k-1)},\mathbf{\Phi},\mathbf{{y}}) \\
&=\mathbf{\hat{x}}^{(k-1)} - \rho^{(k)} \mathbf{\Phi}^{\top} (\mathbf{\Phi} \mathbf{\hat{x}}^{(k-1)} - \mathbf{{y}}) .
\end{aligned}
\end{equation}

\textbf{Controllable Proximal Mapping Module (CPMM):}
Different from ISTA-Net$^+$ \cite{zhang2018ista}, we propose a more effective and robust CPMM to solve the proximal mapping problem in Eq.~(\ref{eq: x2}) with learnable $\psi(\cdot)$. In addition to $\mathbf{r}^{(k)}$ from GDM, CPMM also takes a condition vector $\mathbf{z}=[\gamma, \sigma]$ as input, where $\gamma$ is CS ratio and $\sigma$ is the standard deviation of additive white Gaussian noise in Eq.~(\ref{eq: CS}). In this paper,  $\gamma \in \{10\%, 20\%, 30\%, 40\%, 50\%\}$ and $\sigma \in [0,1]$. That is to say, CPMM can dynamically modulate the network features according to the given $\gamma$ and $\sigma$. Concretely, as shown in Fig.~\ref{fig: opticsframework}, CPMM in the $k$-th phase, denoted as $\mathcal{H}^{(k)}_\text{CPMM}$, is composed of $N_C$ controllable proximal mapping blocks (CPMBs), two convolution layers, and a long skip connection, which can be formulated as follows:
\begin{equation}
\begin{aligned}
& \mathbf{\hat{x}}^{(k)} =\mathcal{H}^{(k)}_\text{CPMM}(\mathbf{r}^{(k)}, \mathbf{z}) \\
& = \mathbf{r}^{(k)}+ \mathcal{W}^{(k)}_{\text{2}}(\mathcal{H}^{(k)}_{\text{CPMB},{N_C}}(...\mathcal{H}_{\text{CPMB},1}^{(k)}(\mathcal{W}_{\text{1}}^{(k)}(\mathbf{r}^{(k)}), \mathbf{z}))),
\end{aligned}
\end{equation}
where the $\mathcal{W}^{(k)}_{\text{1}}$ and $\mathcal{W}^{(k)}_{\text{2}}$ denote the first and the last convolution operations, respectively. Besides, the $j$-th controllable proximal mapping blocks (CPMB) in $\mathcal{H}^{(k)}_\text{CPMM}$ is denoted as $\mathcal{H}^{(k)}_{\text{CPMB},j}$. In our implementation, we set $N_C=3$ by default.

Due to its simplicity and effectiveness, the standard residual block (RB) is adopted as the building block for CPMB, which consists of two convolution layers and a ReLU activation layer. 
In particular, as illustrated in Fig.~\ref{fig:PMB}, each CPMB is composed of a controllable unit (CU) and a residual block (RB), wherein the output of CU is used to modulate the output of the last convolution layer in RB. Each CU, denoted by $\mathcal{H}_\text{CU}$ is efficiently implemented by a fully-connected layer, and takes the condition vector $\mathbf{z}$ as input to generate a controllable 
vector $\mathcal{H}_\text{CU}(\mathbf{z})$ with dimension $C$, which can be defined as:
\begin{align}
     \mathcal{H}_\text{CU}(\mathbf{z}) = W_\text{CU}(\mathbf{z}),
\end{align}
where $W_\text{CU}$ is a fully-connected layer. Furthermore, in our preliminary experiments, it is found that the parameter sharing of $W_\text{CU}$ across all the CUs does not impair the performance. Therefore, we utilized one $W_\text{CU}$ for our COAST to reduce the number of parameters. With $\mathcal{H}_\text{CU}(\mathbf{z})$, the process of CPMB can be formulated as:
\begin{equation}
\begin{aligned}
& \mathbf{F}^{(k)}_{j}  = \mathcal{H}^{(k)}_{\text{CPMB},{j}}(\mathbf{F}^{(k)}_{j-1},\mathbf{z}) \\ 
&= \mathcal{W}^{(k)}_{j,2}(\text{ReLU}(\mathcal{W}^{(k)}_{j,1}(\mathbf{F}^{(k)}_{j-1}))) \cdot \mathcal{H}_\text{CU}(\mathbf{z}) + \mathbf{F}^{(k)}_{j-1} ,
\end{aligned}
\end{equation}
where $\mathbf{F}^{(k)}_{j-1}$ and $\mathbf{F}^{(k)}_j$ are the input and output of $\mathcal{H}^{(k)}_{\text{CPMB},{j}}$, $\text{ReLU}(\cdot)$ denotes a rectified linear unit (ReLU) function, and $\mathcal{W}^{(k)}_{j,1}$ and $\mathcal{W}^{(k)}_{j,2}$ denote the first and last convolution layer of $\mathcal{H}^{(k)}_{\text{CPMB},{j}}$, corresponding to $C$ filters.  $\mathcal{H}_\text{CU}(\mathbf{z})$ is a $C$-dimensional vector used to modulate each channel of $\mathcal{W}^{(k)}_{j,2}(\text{ReLU}(\mathcal{W}^{(k)}_{j,1}(\mathbf{F}^{(k)}_{j-1})))$. In our experimental implementation, we set $C=32$ by default.

\textbf{Plug-and-Play Deblocking (PnP-D):}
To address the issue of blocking artifacts, we propose a plug-and-play deblocking (PnP-D) strategy at testing stage, which is composed of two parts: plug-and-play folding (PnP-F) and plug-and-play unfolding (PnP-U). To be concrete, each image is partitioned into $N_B$ non-overlapped image patches.
As illustrated in  Fig.~\ref{fig: ppd}(a), without PnP-D, each patch is independently processed by the network and its network feature with dimension $N_B \times C \times \sqrt{N} \times \sqrt{N}$ also goes through CPMM independently. However, with PnP-D, in each phase, our COAST first uses PnP-F to  combine $N_B$  sliding  local patches into a whole image's feature with dimension $1 \times C \times  \sqrt{N_BN} \times \sqrt{N_BN}$. Then, the  fully-convolutional CPMM operates on this folded image's feature by exploiting the inter-block relationship in feature domain to effectively eliminate the blocking artifacts. From the output of CPMM, PnP-U is to further extract $N_B$ sliding  local  patches  with dimension $N_B \times C \times \sqrt{N} \times \sqrt{N}$. Obviously, PnP-D is only exploited at the testing stage and enables our COAST to sample independently and reconstruct jointly without introducing any extra parameters. Thus, PnP-D can be used as a plugin and easily embedded into a wide range of existing CS systems without changing the original model structures. Experiments will demonstrate the effectiveness and generalization capability of our PnP-D strategy.

\begin{table*}[t]
\centering
\caption{Average PSNR/SSIM performance comparisons of various CS methods on FRGM with different CS ratios on the Set11 and BSD68 datasets. The best and second best results are highlighted in \textcolor[rgb]{1.00,0.00,0.00}{red} and \textcolor[rgb]{0.00,0.00,1.00}{blue} colors, respectively.}
\label{tab:result_noise_0}
\setlength{\tabcolsep}{5pt}    
\begin{tabular}{cccccccccc} 
\shline
\multicolumn{1}{c|}{\multirow{2}{*}{Datasets}}& \multicolumn{1}{c|}{\multirow{2}{*}{Methods}} & \multicolumn{1}{c|}{\multirow{2}{*}{Category}} &  \multicolumn{5}{c|}{CS Ratio $\gamma$} & \multicolumn{1}{c|}{\multirow{2}{*}{\begin{tabular}[c]{@{}c@{}}Time\\ CPU/GPU \end{tabular}}} & \multicolumn{1}{c}{\multirow{2}{*}{\begin{tabular}[c]{@{}c@{}}FPS\\ CPU/GPU \end{tabular}}} \\ \cline{4-8}
\multicolumn{1}{c|}{} & \multicolumn{1}{c|}{}& \multicolumn{1}{c|}{} & \multicolumn{1}{c|}{10\%}         & \multicolumn{1}{c|}{20\%}         & \multicolumn{1}{c|}{30\%}         & \multicolumn{1}{c|}{40\%}         & \multicolumn{1}{c|}{50\%} & \multicolumn{1}{c|}{}   & \multicolumn{1}{c}{}      \\ \shline

\multicolumn{1}{c}{\multirow{10}{*}{Set11}} & \multicolumn{1}{|l|}{LDAMP \cite{DBLP:conf/nips/MetzlerMB17}}      & \multicolumn{1}{c|}{\multirow{5}{*}{$\mathbf{\Phi}$-Specific}}                   & \multicolumn{1}{c|}{24.71/0.4333} & \multicolumn{1}{c|}{30.65/0.6823} & \multicolumn{1}{c|}{33.87/0.7763} & \multicolumn{1}{c|}{36.03/0.8191} & \multicolumn{1}{c|}{36.60/0.8391} & \multicolumn{1}{c|}{536.6s/-----} & \multicolumn{1}{c}{0.002/-----}\\
\multicolumn{1}{c}{} & \multicolumn{1}{|l|}{ReconNet \cite{lohit2018convolutional}}  & \multicolumn{1}{c|}{}                          & \multicolumn{1}{c|}{24.06/0.7223} & \multicolumn{1}{c|}{26.68/0.8085} & \multicolumn{1}{c|}{28.14/0.8472} & \multicolumn{1}{c|}{30.78/0.8932} & \multicolumn{1}{c|}{31.48/0.8993} & \multicolumn{1}{c|}{----/0.0054s} & \multicolumn{1}{c}{-----/185.2}\\
\multicolumn{1}{c}{} & \multicolumn{1}{|l|}{DPDNN \cite{DBLP:journals/pami/DongWYSWL19}}     & \multicolumn{1}{c|}{}                          & \multicolumn{1}{c|}{24.97/0.7622} & \multicolumn{1}{c|}{28.97/0.8713} & \multicolumn{1}{c|}{32.16/0.9240} & \multicolumn{1}{c|}{34.17/0.9457} & \multicolumn{1}{c|}{36.45/0.9633} & \multicolumn{1}{c|}{----/0.0809s} & \multicolumn{1}{c}{-----/12.36}\\
\multicolumn{1}{c}{} & \multicolumn{1}{|l|}{GDN \cite{gilton2019neumann}}       & \multicolumn{1}{c|}{}                          & \multicolumn{1}{c|}{25.63/0.7782} & \multicolumn{1}{c|}{29.70/0.8785} & \multicolumn{1}{c|}{32.22/0.9199} & \multicolumn{1}{c|}{34.61/0.9461} & \multicolumn{1}{c|}{36.74/0.9629} & \multicolumn{1}{c|}{----/0.0137s} & \multicolumn{1}{c}{-----/72.99}\\
\multicolumn{1}{c}{} & \multicolumn{1}{|l|}{ISTA-Net$^+$ \cite{zhang2018ista}}  & \multicolumn{1}{c|}{}                          & \multicolumn{1}{c|}{26.57/0.8095} & \multicolumn{1}{c|}{30.85/0.9011} & \multicolumn{1}{c|}{33.74/0.9386} & \multicolumn{1}{c|}{\color{blue}36.05/0.9581} & \multicolumn{1}{c|}{\color{blue}38.05/0.9704} & \multicolumn{1}{c|}{----/0.0129s} & \multicolumn{1}{c}{-----/77.52}\\
\cline{2-10}

\multicolumn{1}{c}{} & \multicolumn{1}{|l|}{TVAL3 \cite{DBLP:conf/nips/MetzlerMB17}}     & \multicolumn{1}{c|}{\multirow{5}{*}{$\mathbf{\Phi}$-Agnostic}} & \multicolumn{1}{c|}{24.27/0.7266} & \multicolumn{1}{c|}{27.48/0.8270} & \multicolumn{1}{c|}{29.81/0.8804} & \multicolumn{1}{c|}{31.89/0.9144} & \multicolumn{1}{c|}{33.97/0.9401} & \multicolumn{1}{c|}{5.438s/-----} & \multicolumn{1}{c}{0.184/-----}  \\
\multicolumn{1}{c}{} & \multicolumn{1}{|l|}{BM3D-AMP \cite{metzler2016denoising}}  & \multicolumn{1}{c|}{}                          & \multicolumn{1}{c|}{22.60/0.3489} & \multicolumn{1}{c|}{26.77/0.5456} & \multicolumn{1}{c|}{30.26/0.6580} & \multicolumn{1}{c|}{33.66/0.7316} & \multicolumn{1}{c|}{35.93/0.7792} & \multicolumn{1}{c|}{62.08s/-----} & \multicolumn{1}{c}{0.016/-----}\\
\multicolumn{1}{c}{} & \multicolumn{1}{|l|}{DIP \cite{ulyanov2018deep}}       & \multicolumn{1}{c|}{}                          & \multicolumn{1}{c|}{25.98/0.7619} & \multicolumn{1}{c|}{29.81/0.8683} & \multicolumn{1}{c|}{33.25/0.9330} & \multicolumn{1}{c|}{33.41/0.8933} & \multicolumn{1}{c|}{35.96/0.9480} & \multicolumn{1}{c|}{----/491.30s} & \multicolumn{1}{c}{-----/0.002} \\
\multicolumn{1}{c}{} & \multicolumn{1}{|l|}{NLR-CSNet \cite{8999514}}   & \multicolumn{1}{c|}{}                          & \multicolumn{1}{c|}{\color{blue}28.05/0.8477} & \multicolumn{1}{c|}{\color{blue}31.64/0.9126} & \multicolumn{1}{c|}{\color{blue}33.89/0.9391} & \multicolumn{1}{c|}{35.65/0.9545} & \multicolumn{1}{c|}{37.12/0.9635} & \multicolumn{1}{c|}{----/572.95s} & \multicolumn{1}{c}{-----/0.002} \\
\multicolumn{1}{c}{} & \multicolumn{1}{|l|}{COAST}    &  \multicolumn{1}{c|}{}    & \multicolumn{1}{c|}{\color{red}28.69/0.8618} & \multicolumn{1}{c|}{\color{red}32.54/0.9251} & \multicolumn{1}{c|}{\color{red}35.04/0.9501} & \multicolumn{1}{c|}{\color{red}37.13/0.9648} & \multicolumn{1}{c|}{\color{red}38.94/0.9744} & \multicolumn{1}{c|}{----/0.0248s} & \multicolumn{1}{c}{-----/40.32} \\ \hline 

\multicolumn{1}{c}{\multirow{10}{*}{BSD68}} & \multicolumn{1}{|l|}{LDAMP \cite{DBLP:conf/nips/MetzlerMB17}}     & \multicolumn{1}{c|}{\multirow{5}{*}{$\mathbf{\Phi}$-Specific}}                   & \multicolumn{1}{c|}{23.94/0.3175} & \multicolumn{1}{c|}{27.74/0.5762} & \multicolumn{1}{c|}{\color{blue}30.28/0.7160} & \multicolumn{1}{c|}{\color{blue}32.12/0.7819} & \multicolumn{1}{c|}{32.89/0.7863} & \multicolumn{1}{c|}{776.0s/-----} & \multicolumn{1}{c}{0.001/-----}\\
\multicolumn{1}{c}{} & \multicolumn{1}{|l|}{ReconNet \cite{lohit2018convolutional}}  & \multicolumn{1}{c|}{}                          & \multicolumn{1}{c|}{23.88/0.6400} & \multicolumn{1}{c|}{25.75/0.7317} & \multicolumn{1}{c|}{26.72/0.7870} & \multicolumn{1}{c|}{28.96/0.8499} & \multicolumn{1}{c|}{30.13/0.8798} & \multicolumn{1}{c|}{----/0.0053s} & \multicolumn{1}{c}{-----/188.7} \\
\multicolumn{1}{c}{} & \multicolumn{1}{|l|}{DPDNN \cite{DBLP:journals/pami/DongWYSWL19}}     & \multicolumn{1}{c|}{}                          & \multicolumn{1}{c|}{24.26/0.6706} & \multicolumn{1}{c|}{27.02/0.7893} & \multicolumn{1}{c|}{29.32/0.8615} & \multicolumn{1}{c|}{31.06/0.9011} & \multicolumn{1}{c|}{32.97/0.9324} & \multicolumn{1}{c|}{----/0.1293s} & \multicolumn{1}{c}{-----/7.734}  \\
\multicolumn{1}{c}{} & \multicolumn{1}{|l|}{GDN \cite{gilton2019neumann}}       & \multicolumn{1}{c|}{}                          & \multicolumn{1}{c|}{24.86/0.6827} & \multicolumn{1}{c|}{27.54/0.7963} & \multicolumn{1}{c|}{29.52/0.8602} & \multicolumn{1}{c|}{31.49/0.9047} & \multicolumn{1}{c|}{33.35/0.9351} & \multicolumn{1}{c|}{----/0.0127s} & \multicolumn{1}{c}{-----/78.74}  \\
\multicolumn{1}{c}{} & \multicolumn{1}{|l|}{ISTA-Net$^+$ \cite{zhang2018ista}} & \multicolumn{1}{c|}{}                          & \multicolumn{1}{c|}{\color{blue}25.24/0.6991} & \multicolumn{1}{c|}{\color{blue}28.00/0.8142} & \multicolumn{1}{c|}{30.20/0.8771} & \multicolumn{1}{c|}{32.10/0.9155} & \multicolumn{1}{c|}{\color{blue}33.93/0.9421} & \multicolumn{1}{c|}{----/0.0118s} & \multicolumn{1}{c}{-----/84.75} \\
\cline{2-10} 

\multicolumn{1}{c}{} & \multicolumn{1}{|l|}{TVAL3 \cite{DBLP:conf/nips/MetzlerMB17}}     & \multicolumn{1}{c|}{\multirow{5}{*}{$\mathbf{\Phi}$-Agnostic}}           & \multicolumn{1}{c|}{24.29/0.6507} & \multicolumn{1}{c|}{26.44/0.7483} & \multicolumn{1}{c|}{28.12/0.8101} & \multicolumn{1}{c|}{29.91/0.8658} & \multicolumn{1}{c|}{31.62/0.9050} & \multicolumn{1}{c|}{7.096s/-----} & \multicolumn{1}{c}{0.141/-----} \\
\multicolumn{1}{c}{} & \multicolumn{1}{|l|}{BM3D-AMP \cite{metzler2016denoising}}  & \multicolumn{1}{c|}{}                          & \multicolumn{1}{c|}{22.68/0.2484} & \multicolumn{1}{c|}{24.77/0.3974} & \multicolumn{1}{c|}{26.44/0.5031} & \multicolumn{1}{c|}{28.19/0.5955} & \multicolumn{1}{c|}{29.86/0.6855} & \multicolumn{1}{c|}{90.40s/-----} & \multicolumn{1}{c}{0.011/-----}  \\
\multicolumn{1}{c}{} & \multicolumn{1}{|l|}{DIP \cite{ulyanov2018deep}}       & \multicolumn{1}{c|}{}                          & \multicolumn{1}{c|}{25.05/0.6932} & \multicolumn{1}{c|}{27.25/0.7823} & \multicolumn{1}{c|}{28.66/0.8248} & \multicolumn{1}{c|}{29.82/0.8430} & \multicolumn{1}{c|}{31.21/0.8689} & \multicolumn{1}{c|}{----/642.30s} & \multicolumn{1}{c}{-----/0.002}\\
\multicolumn{1}{c}{} & \multicolumn{1}{|l|}{NLR-CSNet \cite{8999514}} & \multicolumn{1}{c|}{}                          & \multicolumn{1}{c|}{25.23/0.7131} & \multicolumn{1}{c|}{27.69/0.8104} & \multicolumn{1}{c|}{29.55/0.8649} & \multicolumn{1}{c|}{31.14/0.9004} & \multicolumn{1}{c|}{32.57/0.9243} & \multicolumn{1}{c|}{----/788.99s} & \multicolumn{1}{c}{-----/0.001}  \\
\multicolumn{1}{c}{} & \multicolumn{1}{|l|}{COAST}    & \multicolumn{1}{c|}{}          & \multicolumn{1}{c|}{\color{red} 26.28/0.7422} & \multicolumn{1}{c|}{\color{red} 29.00/0.8413} & \multicolumn{1}{c|}{\color{red} 31.06/0.8934} & \multicolumn{1}{c|}{\color{red} 32.93/0.9267} & \multicolumn{1}{c|}{\color{red} 34.74/0.9497} & \multicolumn{1}{c|}{----/0.0276s} & \multicolumn{1}{c}{-----/36.23}  \\ \shline
\end{tabular}
\end{table*}

\subsection{Network Parameters and Loss Function}
\label{sec:loss}
Given $L (L\geq 1)$ sampling matrices with different dimensions $M_l$ by $N_l$ $(l\leq L)$, we first utilize RPA to generate the augmented sampling matrix set $\{\mathbf{\Phi}_t\}_{t=1}^{N_\Phi}, N_\Phi=L\times N_S$. Next, given the training dataset $\left \{ (\mathbf{x}_i) \right \}_{i=1}^{N_D}$, we get $\mathbf{y}_{i,t}=\mathbf{\Phi}_{t}(\mathbf{x}_{i}+\mathbf{n}_i), \mathbf{n}_i\sim \mathcal{N}(0,\sigma_i^{2})$. Then, with $\mathbf{z}_i=[\gamma_i, \sigma_i]$ and taking $\mathbf{y}_{i,t}, \mathbf{\Phi}_t, \mathbf{z}_i$ as input, COAST aims to reduce the discrepancy between $\mathbf{x}_i$ and $\mathcal{H}_{\text{COAST}}(\mathbf{y}_{i,t}, \mathbf{\Phi}_t,\mathbf{z}_i)$. Therefore, we design the end-to-end loss function for our COAST as follows:
\begin{equation}
\label{eq: LossFunction}
{\mathcal{L}}(\mathbf{\Theta}) =  \frac{{\sum^{N_D}_{i=1}\sum^{N_\Phi}_{t=1}\|\mathcal{H}_{\text{COAST}}(\mathbf{y}_{i,t},\mathbf{\Phi}_t,\mathbf{z}_i) - \mathbf{x}_i\|^2_2}}{N_DNN_\Phi},
\end{equation}
where $N_D$ denotes the total number of training patches of size $\sqrt{N}$$\times$$\sqrt{N}$. And $\mathbf{\Theta}$ denotes the learnable parameter set in COAST, includes the parameters of the GDM and CPMM $\mathcal{H}^{(k)}_\text{GDM}(\cdot), \mathcal{H}^{(k)}_\text{CPMM}(\cdot)$ in the recovery subnet. As such, $\mathbf{\Theta} = \{ \mathcal{H}^{(k)}_\text{GDM}(\cdot), \mathcal{H}^{(k)}_\text{CPMM}(\cdot)\}^{N_P}_{k=1}$.

At the testing stage, the users are allowed to use an arbitrary sampling matrix $\mathbf{\Phi} \in \mathbb{R}^{M\times N}$ (either $\mathbf{\Phi} \in \{\mathbf{\Phi}_t\}$ or $\mathbf{\Phi} \notin \{\mathbf{\Phi}_t\}$), which is used to obtain a CS measurement $\mathbf{y = \Phi (x + n)}$, where $\gamma=\frac{M}{N}$ and $\mathbf{n} \sim \mathcal{N}(0,\sigma^{2})$. Then, given $ \mathbf{y}, \mathbf{\Phi}, \mathbf{z}=[\gamma, \sigma]$ as input, our COAST can generate a desirable reconstruction result by $\mathcal{H}_{\text{COAST}}(\mathbf{y},\mathbf{\Phi},\mathbf{z})$.



\section{Experiments}

\subsection{Implementation Details}

Following common practice \cite{kulkarni2016reconnet,zhang2018ista} and for fair comparison, we employ the same set of 91 images provided in \cite{DBLP:conf/eccv/DongLHT14} in the training stage and extract the luminance component of $88912$ randomly cropped image patches (each of default size $33\times 33$), \textit{i.e.}, $N_D=88912$ and $N=1089$.  

Our COAST is implemented in PyTorch \cite{pytorch} and is trained on a workstation with Intel Core i7-6820 CPU and GTX1080Ti GPU with a batch size 64. We use Adam \cite{kingma2015adam}, as well as momentum of 0.9 and weight decay of 0.999 for parameter optimization.
Training COAST with phase number $N_P = 20$ roughly takes four days.
For testing, we utilize two widely-used benchmark datasets: Set11 \cite{kulkarni2016reconnet} and BSD68 \cite{martin2001database}. 
Note that we deal with color images in the transformed YCbCr space and the CS recovered results are evaluated with Peak  Signal-to-Noise  Ratio (PSNR)  and  Structural  Similarity  Index (SSIM) \cite{wang2004image} on Y channel (\textit{i.e.}, luminance).
\begin{figure*}[t]
\centering
\includegraphics[width=1.0\textwidth]{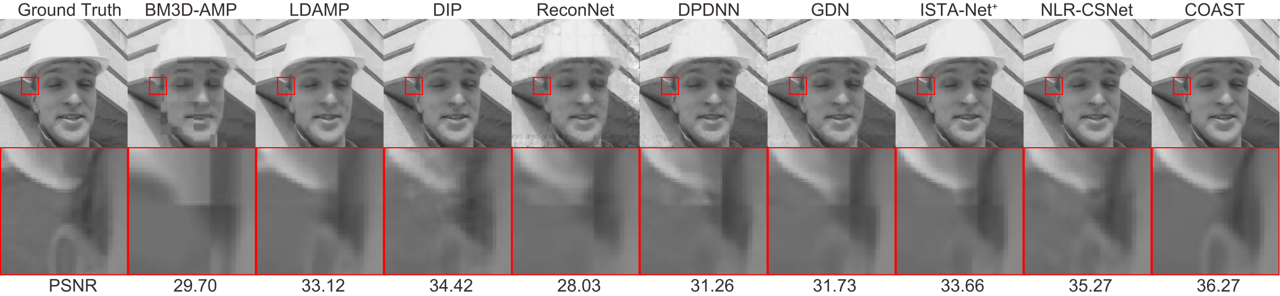}
\caption{Visual comparison on Set11 dataset of various CS methods on fixed random Gaussian matrix (FRGM) in the case of $\gamma = 10\% $.}
\label{fig: set11ratio10fixed}  
\vspace{0pt}
\end{figure*}

\begin{figure*}[t]
\centering
\includegraphics[width=1.0\textwidth]{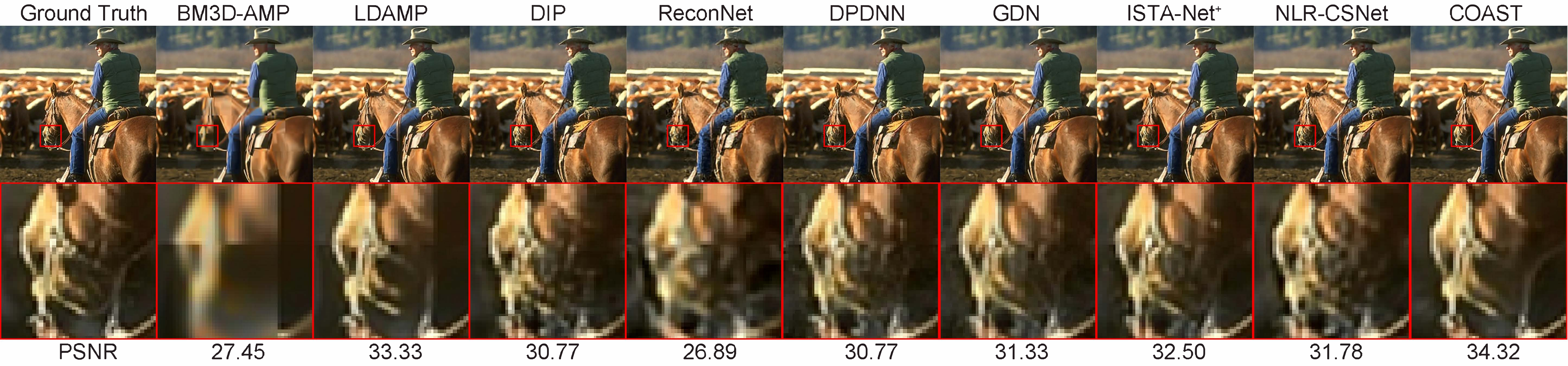}
\caption{Visual comparison on BSD68 dataset of various CS methods on fixed random Gaussian matrix (FRGM) in the case of $\gamma = 30\% $.}
\label{fig: bsd68ratio30fixed}  
\vspace{0pt}
\end{figure*}

\vspace{-5pt}
\subsection{Comparison with State-of-the-Art Methods}
\label{sota}

Our proposed COAST is compared with ten representative state-of-the-art methods including TVAL3 \cite{DBLP:conf/nips/MetzlerMB17}, BM3D-AMP \cite{metzler2016denoising}, LDAMP \cite{DBLP:conf/nips/MetzlerMB17}, DIP \cite{ulyanov2018deep}, ReconNet \cite{lohit2018convolutional}, DPDNN \cite{DBLP:journals/pami/DongWYSWL19}, GDN \cite{gilton2019neumann}, ISTA-Net$^+$ \cite{zhang2018ista}, NLR-CSNet \cite{8999514}, and OPINE-Net \cite{9019857}. 
Here, the acquired measurement is 'noiseless', \textit{i.e.,} the standard deviation $\sigma$ of $\mathbf{n}$ in Eq.~(\ref{eq: CS}) is set to 0. 
To be concrete, TVAL3, BM3D-AMP are traditional optimization-based methods; ReconNet is a classical deep network-based method; GDN, LDAMP, DPDNN, ISTA-Net$^+$, and OPINE-Net are deep unfolding methods; DIP and NLR-CSNet are unsupervised methods based on the idea that the structure of the auto-encoder network itself is a good prior to capture image statistics. Specifically, when handling multiple sampling matrices, ReconNet, GDN, LDAMP, DPDNN, and ISTA-Net$^+$ need to re-train the model with respect to each sampling matrix, while TVAL3, BM3D-AMP, DIP, and NLR-CSNet need to solve the computationally expensive optimization with hand-crafted prior or deep image prior for each image. However, our COAST can handle arbitrary sampling matrices with a single trained model, leading to high performance and fast speed. According to whether a 
network model needs to be re-trained or not for various sampling matrices, we generally divide CS methods into two categories: $\mathbf{\Phi}$-\textbf{specific} and $\mathbf{\Phi}$-\textbf{agnostic}. In the following, three comparisons with different sampling matrices are provided, \textit{i.e.,} fixed random Gaussian matrix (\textbf{FRGM}), data-driven adaptively learned matrix (\textbf{DALM}) and unseen sampling matrix (\textbf{USM}).

\begin{table}[t]
\centering

\caption{Average PSNR performance comparisons of various CS methods on DALM with different CS ratios on the Set11 and BSD68 datasets. The best and second best results are highlighted in \textcolor[rgb]{1.00,0.00,0.00}{red} and \textcolor[rgb]{0.00,0.00,1.00}{blue} colors, respectively.}
\label{tab:result_datadriven_set11}
\setlength{\tabcolsep}{4pt}
\begin{tabular}{cccccc}
\shline
\multicolumn{1}{c|}{\multirow{2}{*}{Datasets}} & \multicolumn{1}{c|}{\multirow{2}{*}{Methods}}   & \multicolumn{1}{c|}{\multirow{2}{*}{Category}}  & \multicolumn{3}{c}{CS Ratio $\gamma$} \\  \cline{4-6}
\multicolumn{1}{c|}{}  & \multicolumn{1}{c|}{} & \multicolumn{1}{c|}{}  &\multicolumn{1}{c|}{10\%}         & \multicolumn{1}{c|}{30\%}         & \multicolumn{1}{c}{50\%}            \\ \shline

\multicolumn{1}{c}{\multirow{13}{*}{Set11}}  &\multicolumn{1}{|l|}{LDAMP \cite{DBLP:conf/nips/MetzlerMB17}}      & \multicolumn{1}{c|}{\multirow{8}{*}{$\mathbf{\Phi}$-Specific}}                    & \multicolumn{1}{c|}{23.24} & \multicolumn{1}{c|}{29.40} & \multicolumn{1}{c}{31.64} \\
\multicolumn{1}{c}{} &\multicolumn{1}{|l|}{ReconNet \cite{lohit2018convolutional}}  & \multicolumn{1}{c|}{}                          & \multicolumn{1}{c|}{26.97} & \multicolumn{1}{c|}{32.04} & \multicolumn{1}{c}{34.31} \\
\multicolumn{1}{c}{} &\multicolumn{1}{|l|}{DPDNN \cite{DBLP:journals/pami/DongWYSWL19}}     & \multicolumn{1}{c|}{}                    & \multicolumn{1}{c|}{27.74} & \multicolumn{1}{c|}{34.29} & \multicolumn{1}{c}{38.80} \\
\multicolumn{1}{c}{} &\multicolumn{1}{|l|}{GDN \cite{gilton2019neumann}}       & \multicolumn{1}{c|}{}                      & \multicolumn{1}{c|}{28.77} & \multicolumn{1}{c|}{35.01} & \multicolumn{1}{c}{39.14} \\
\multicolumn{1}{c}{} &\multicolumn{1}{|l|}{ISTA-Net$^+$ \cite{zhang2018ista}}  & \multicolumn{1}{c|}{}                      & \multicolumn{1}{c|}{29.10} & \multicolumn{1}{c|}{35.45} & \multicolumn{1}{c}{39.84} \\
\multicolumn{1}{c}{}
&\multicolumn{1}{|l|}{CSNet \cite{shi2017deep}}  & \multicolumn{1}{c|}{}  & \multicolumn{1}{c|}{28.10} & \multicolumn{1}{c|}{33.86} & \multicolumn{1}{c}{37.51} \\
\multicolumn{1}{c}{}
&\multicolumn{1}{|l|}{SCSNet \cite{shi2019scalable}}   & \multicolumn{1}{c|}{}  & \multicolumn{1}{c|}{28.48} & \multicolumn{1}{c|}{34.62} & \multicolumn{1}{c}{39.01} \\
\multicolumn{1}{c}{} &\multicolumn{1}{|l|}{OPINE-Net \cite{9019857}}   & \multicolumn{1}{c|}{}  & \multicolumn{1}{c|}{29.71} & \multicolumn{1}{c|}{\color{blue}35.92} &  \multicolumn{1}{c}{\color{blue}40.12} \\ \cline{2-6}

\multicolumn{1}{c}{} & \multicolumn{1}{|l|}{TVAL3 \cite{DBLP:conf/nips/MetzlerMB17}}     & \multicolumn{1}{c|}{\multirow{5}{*}{$\mathbf{\Phi}$-Agnostic}}     & \multicolumn{1}{c|}{25.83} & \multicolumn{1}{c|}{31.06} & \multicolumn{1}{c}{35.45} \\
\multicolumn{1}{c}{} &\multicolumn{1}{|l|}{BM3D-AMP \cite{metzler2016denoising}}  & \multicolumn{1}{c|}{}                          & \multicolumn{1}{c|}{27.41} & \multicolumn{1}{c|}{33.60} & \multicolumn{1}{c}{37.64} \\
\multicolumn{1}{c}{} &\multicolumn{1}{|l|}{DIP \cite{ulyanov2018deep}}       & \multicolumn{1}{c|}{}                          & \multicolumn{1}{c|}{28.97} & \multicolumn{1}{c|}{33.23} & \multicolumn{1}{c}{38.25} \\
\multicolumn{1}{c}{} &\multicolumn{1}{|l|}{NLR-CSNet \cite{8999514}}   & \multicolumn{1}{c|}{}  & \multicolumn{1}{c|}{\color{blue}29.80} & \multicolumn{1}{c|}{35.29} & \multicolumn{1}{c}{38.93} \\
\multicolumn{1}{c}{}
&\multicolumn{1}{|l|}{COAST}    & \multicolumn{1}{c|}{}      & \multicolumn{1}{c|}{\color{red}30.03} & \multicolumn{1}{c|}{\color{red} 36.35} & \multicolumn{1}{c}{\color{red} 40.32} \\ \hline 

\multicolumn{1}{c}{\multirow{13}{*}{BSD68}} &\multicolumn{1}{|l|}{LDAMP \cite{DBLP:conf/nips/MetzlerMB17}}     & \multicolumn{1}{c|}{\multirow{8}{*}{$\mathbf{\Phi}$-Specific}}                           & \multicolumn{1}{c|}{22.39} & \multicolumn{1}{c|}{26.59} & \multicolumn{1}{c}{29.73} \\
\multicolumn{1}{c}{} &\multicolumn{1}{|l|}{ReconNet \cite{lohit2018convolutional}}  & \multicolumn{1}{c|}{}                          & \multicolumn{1}{c|}{26.16} & \multicolumn{1}{c|}{30.00} & \multicolumn{1}{c}{31.88} \\
\multicolumn{1}{c}{} &\multicolumn{1}{|l|}{DPDNN \cite{DBLP:journals/pami/DongWYSWL19}}     & \multicolumn{1}{c|}{}                          & \multicolumn{1}{c|}{26.49} & \multicolumn{1}{c|}{31.24} & \multicolumn{1}{c}{35.14} \\
\multicolumn{1}{c}{} &\multicolumn{1}{|l|}{GDN \cite{gilton2019neumann}}       & \multicolumn{1}{c|}{}                          & \multicolumn{1}{c|}{27.15} & \multicolumn{1}{c|}{31.80} & \multicolumn{1}{c}{35.52} \\
\multicolumn{1}{c}{} &\multicolumn{1}{|l|}{ISTA-Net$^+$ \cite{zhang2018ista}} & \multicolumn{1}{c|}{}                          & \multicolumn{1}{c|}{27.27} & \multicolumn{1}{c|}{32.02} & \multicolumn{1}{c}{35.90} \\
\multicolumn{1}{c}{}
&\multicolumn{1}{|l|}{CSNet \cite{shi2017deep}}  & \multicolumn{1}{c|}{}  & \multicolumn{1}{c|}{27.10} & \multicolumn{1}{c|}{31.45} & \multicolumn{1}{c}{34.89} \\
\multicolumn{1}{c}{}
&\multicolumn{1}{|l|}{SCSNet \cite{shi2019scalable}}   & \multicolumn{1}{c|}{}  & \multicolumn{1}{c|}{27.28} & \multicolumn{1}{c|}{31.87} & \multicolumn{1}{c}{35.77} \\
\multicolumn{1}{c}{} &\multicolumn{1}{|l|}{OPINE-Net \cite{9019857}}   & \multicolumn{1}{c|}{}  & \multicolumn{1}{c|}{\color{blue}27.61} & \multicolumn{1}{c|}{\color{blue}32.35} & \multicolumn{1}{c}{\color{blue}36.11} \\ \cline{2-6}

\multicolumn{1}{c}{} & \multicolumn{1}{|l|}{TVAL3 \cite{DBLP:conf/nips/MetzlerMB17}}     & \multicolumn{1}{c|}{\multirow{5}{*}{$\mathbf{\Phi}$-Agnostic}} & \multicolumn{1}{c|}{25.63} & \multicolumn{1}{c|}{29.68} & \multicolumn{1}{c}{33.24} \\
\multicolumn{1}{c}{} &\multicolumn{1}{|l|}{BM3D-AMP \cite{metzler2016denoising}}  & \multicolumn{1}{c|}{}                          & \multicolumn{1}{c|}{25.68} & \multicolumn{1}{c|}{28.89} & \multicolumn{1}{c}{31.86} \\
\multicolumn{1}{c}{} &\multicolumn{1}{|l|}{DIP \cite{ulyanov2018deep}}       & \multicolumn{1}{c|}{}                          & \multicolumn{1}{c|}{26.74} & \multicolumn{1}{c|}{30.33} & \multicolumn{1}{c}{33.06} \\
\multicolumn{1}{c}{} &\multicolumn{1}{|l|}{NLR-CSNet \cite{8999514}} & \multicolumn{1}{c|}{}                          & \multicolumn{1}{c|}{27.33} & \multicolumn{1}{c|}{31.36} & \multicolumn{1}{c}{34.67} \\
\multicolumn{1}{c}{} &\multicolumn{1}{|l|}{COAST}    & \multicolumn{1}{c|}{}     & \multicolumn{1}{c|}{\color{red}27.77} & \multicolumn{1}{c|}{\color{red}32.56} & \multicolumn{1}{c}{\color{red}36.36} \\ \shline
\end{tabular}
\vspace{-10pt}
\end{table}

\textbf{Comparison on FRGM:} For FRGM, to demonstrate the effectiveness of our approach, we utilize the same five specific sampling matrices as \cite{zhang2018ista}, which are separately denoted by 
$\mathbf{\Phi}_{109 \times 1089}$, $\mathbf{\Phi}_{218 \times 1089}$, $\mathbf{\Phi}_{327 \times 1089}$, $\mathbf{\Phi}_{436 \times 1089}$, $\mathbf{\Phi}_{545 \times 1089}$, corresponding to CS ratio $\gamma = 10\%~(M_1 = 109, N = 1089), \gamma = 20\%~(M_2 = 218, N = 1089), \gamma = 30\%~(M_3 = 327, N = 1089), \gamma = 40\%~(M_4 = 436, N = 1089), \gamma = 50\%~(M_5 = 545, N = 1089$). At the training stage, the $\mathbf{\Phi}$-specific methods need to train  their models separately for each sampling matrix. Instead, for COAST, we generate $N_\Phi = L \times N_{S} = 5 \times 25 = 125$ sampling matrices with $\mathcal{RPA}\left( \left[ \mathbf{\Phi}_{M_1\times N}, \dots, \mathbf{\Phi}_{M_5 \times N} \right] \right)$, leading to the augmented sampling set $\{\mathbf{\Phi}_t\}_{t=1}^{125}$ = $\{\mathbf{\Phi}_{109 \times 1089 \circledast 1}$, $\dots$, $\mathbf{\Phi}_{109 \times 1089 \circledast 25}$, $\mathbf{\Phi}_{218 \times 1089 \circledast 1}$, $\dots$, $\mathbf{\Phi}_{218 \times 1089 \circledast 25}$, $\mathbf{\Phi}_{327 \times 1089 \circledast 1}$, $\dots$, $\mathbf{\Phi}_{327 \times 1089 \circledast 25}$, $\mathbf{\Phi}_{436 \times 1089 \circledast 1}$, $\dots$, $\mathbf{\Phi}_{436 \times 1089 \circledast 25}$, $\mathbf{\Phi}_{545 \times 1089 \circledast 1}$, $\dots$, $\mathbf{\Phi}_{545 \times 1089 \circledast 25}\}$. We exploit $\{\mathbf{\Phi}_t\}_{t=1}^{125}$ to train our COAST only once. 

Table~\ref{tab:result_noise_0} lists the average PSNR and SSIM results for different ratios on Set11 and BSD68 datasets, respectively.
One can observe that, as a deep unfolding method, ISTA-Net$^+$ has a great improvement compared with most of the competing methods, and the recent unsupervised method NLR-CSNet performs better than ISTA-Net$^+$ when the CS ratio is low.
Benefited from the proposed random projection augmentation (RPA) strategy, controllable proximal mapping module (CPMM), and plug-and-play deblocking (PnP-D) strategy, our COAST achieves the highest PSNR/SSIM results.
For example, our COAST achieves on average 1.46/1.54 dB and 1.24/0.88 dB PSNR gains over the state-of-the-art methods NLR-CSNet and ISTA-Net$^{+}$ on Set11/BSD68 dataset, respectively.
The last two columns in Table~\ref{tab:result_noise_0} show the run-time analysis of all the competing methods, which clearly indicates that the proposed COAST produce consistently better reconstruction results, while remaining computationally attractive real-time speed. Moreover, among all the $\mathbf{\Phi}$-agnostic methods, our COAST has the fastest running speed with GPU.
Fig.~\ref{fig: set11ratio10fixed} and Fig.~\ref{fig: bsd68ratio30fixed} further show the visual comparisons of all the competing methods on two test images when CS ratio $\gamma$ is 10\% and 30\% respectively. Obviously, the proposed COAST is able to recover more texture details and much sharper edges than other competing methods. 

\begin{figure*}[t]
\centering
\includegraphics[width=1.0\textwidth]{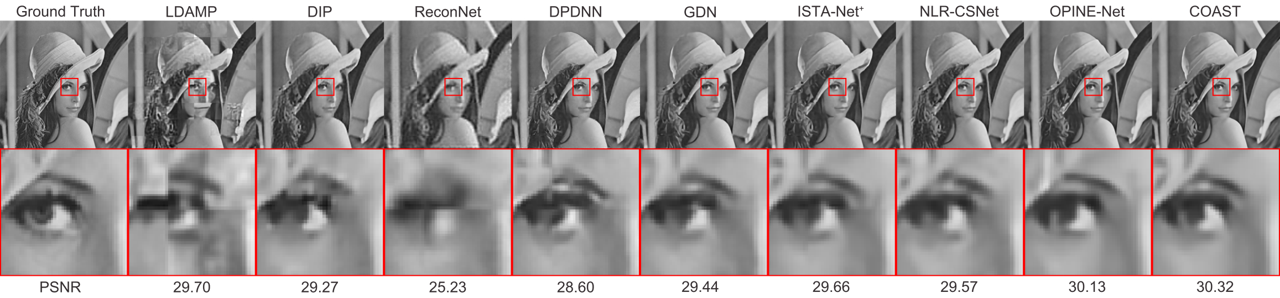}
\caption{Visual comparison on Set11 dataset of various CS methods on data-driven adaptively learned matrix (DALM) in the case of $\gamma = 10\% $.}
\label{fig: set11ratio10data}  
\vspace{0pt}
\end{figure*}
%

\begin{figure*}[t]
\centering
\includegraphics[width=1\textwidth]{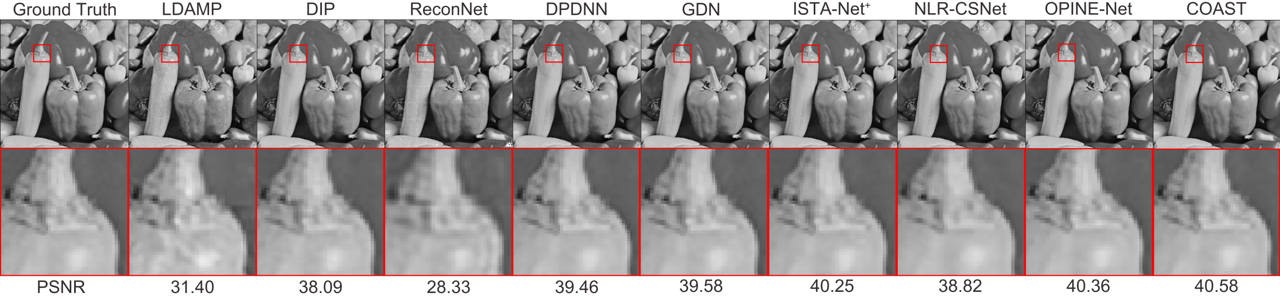}
\caption{Visual comparison on Set11 dataset of various CS methods on data-driven adaptively learned matrix (DALM) in the case of $\gamma = 50\% $.}
\label{fig: set11ratio50data}  
\vspace{0pt}
\end{figure*}

\textbf{Comparison on DALM:} For DALM, three learned sampling matrices produced by OPINE-Net \cite{9019857} are used, which are separately denoted by, 
$\mathbf{\Phi}_{109 \times 1089}, \mathbf{\Phi}_{327 \times 1089}, \mathbf{\Phi}_{545 \times 1089}$, corresponding to CS ratio $\gamma = 10\%~(M_1 = 109, N = 1089), \gamma = 30\%~(M_2 = 327, N = 1089), \gamma = 50\%~(M_3 = 545, N = 1089)$. At training, the $\mathbf{\Phi}$-specific methods train  their models separately for each sampling matrix.   
Similar to FRGM, we adopt
$\mathcal{RPA}\left( \left[ \mathbf{\Phi}_{M_1\times N}, \dots, \mathbf{\Phi}_{M_3 \times N} \right] \right)$ to
generate $N_\Phi = L \times N_S = 3 \times 5 = 15$ augmented sampling matrices to train our COAST only once. 

Table~\ref{tab:result_datadriven_set11} reports the average PSNR and SSIM results of various methods for three CS ratios under the condition of DALM. Comparing Table~\ref{tab:result_datadriven_set11} and  Table~\ref{tab:result_noise_0}, one can observe that the same method with the same CS ratio in the case DALM generally obtains more than 1 dB gains than that in FRGM, which indicates that DALM can preserve more image information than FRGM. Moreover,  Table~\ref{tab:result_datadriven_set11} clearly shows that our COAST not only achieves the best performance among all the competing methods, but also enables dealing with multiple DALMs with a single model, which is in accordance with the results on FRGM. In particular, our COAST improves roughly 0.32 dB and 0.21 dB on average PSNR over three CS ratios, in comparison with the state-of-the-art method OPINE-Net on Set11 and BSD68 datasets, respectively. 
Fig.~\ref{fig: set11ratio10data} and Fig.~\ref{fig: set11ratio50data}  illustrate the visual results of various methods and further demonstrate the effectiveness and superiority of our method.
\begin{figure}[t]
\setlength{\abovecaptionskip}{0.cm}
\setlength{\belowcaptionskip}{1cm}
\centering
\includegraphics[width=1\linewidth]{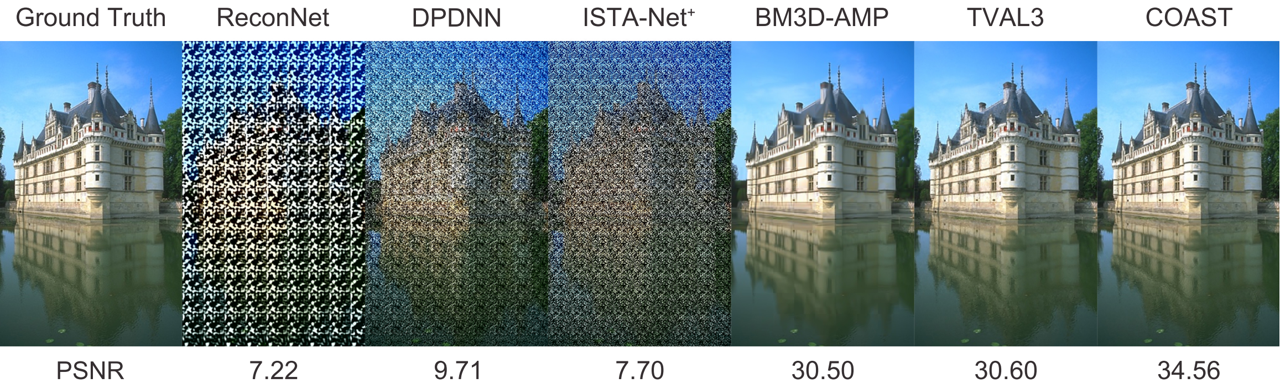} 
\caption{Visual comparison on BSD68 dataset of various CS methods on unseen sampling matrix (USM) in the case of $\gamma = 50\% $.}
\label{fig:unseen50}
\end{figure}
\begin{table*}[t]
\centering
\caption{Average PSNR/SSIM performance comparisons of various CS methods on USM with different CS ratios on the Set11 dataset. The best and second best results are highlighted in \textcolor[rgb]{1.00,0.00,0.00}{red} and \textcolor[rgb]{0.00,0.00,1.00}{blue} colors, respectively.} 
\label{tab:diff_seen_unseen}
\begin{tabular}{ccccccc}
\shline
\multicolumn{1}{c|}{\multirow{2}{*}{Methods}}  & \multicolumn{1}{c|}{\multirow{2}{*}{Category}}  & \multicolumn{5}{c}{CS Ratio $\gamma$} 

\\ \cline{3-7}  \multicolumn{1}{c|}{} & \multicolumn{1}{c|}{} & \multicolumn{1}{c|}{10\%}         & \multicolumn{1}{c|}{20\%}         & \multicolumn{1}{c|}{30\%}         & \multicolumn{1}{c|}{40\%}         & \multicolumn{1}{c}{50\%}         \\ \shline 
\multicolumn{1}{l|}{DPDNN \cite{DBLP:journals/pami/DongWYSWL19}}     & \multicolumn{1}{c|}{\multirow{3}{*}{$\mathbf{\Phi}$-Specific}} &  \multicolumn{1}{c|}{8.25/0.0519} & \multicolumn{1}{c|}{8.44/0.0692} & \multicolumn{1}{c|}{8.88/0.0791} & \multicolumn{1}{c|}{9.20/0.0873} & \multicolumn{1}{c}{8.85/0.0774} \\
\multicolumn{1}{l|}{ReconNet \cite{lohit2018convolutional}}   & \multicolumn{1}{c|}{}                          & \multicolumn{1}{c|}{10.52/0.0434} & \multicolumn{1}{c|}{ 6.18/0.0076} & \multicolumn{1}{c|}{6.83/0.0126} & \multicolumn{1}{c|}{6.36/0.0123} & \multicolumn{1}{c}{6.47/0.0069} \\
\multicolumn{1}{l|}{ISTA-Net$^+$ \cite{zhang2018ista}}  & \multicolumn{1}{c|}{}                          &  \multicolumn{1}{c|}{16.87/0.2708} & \multicolumn{1}{c|}{10.81/0.08551} & \multicolumn{1}{c|}{6.84/0.0259} & \multicolumn{1}{c|}{6.26/0.0153} & \multicolumn{1}{c}{6.42/0.0198} \\ \hline

\multicolumn{1}{l|}{TVAL3 \cite{DBLP:conf/nips/MetzlerMB17}}  & \multicolumn{1}{c|}{\multirow{3}{*}{$\mathbf{\Phi}$-Agnostic}}                         &  \multicolumn{1}{c|}{\color{blue} 24.21/0.7246} & \multicolumn{1}{c|}{\color{blue}27.34/0.8218} & \multicolumn{1}{c|}{29.81/0.8797} & \multicolumn{1}{c|}{31.82/0.9134} & \multicolumn{1}{c}{33.98/0.9401} \\
\multicolumn{1}{l|}{BM3D-AMP \cite{metzler2016denoising}} & \multicolumn{1}{c|}{}  & \multicolumn{1}{c|}{22.47/0.3409} & \multicolumn{1}{c|}{26.80/0.5457} & \multicolumn{1}{c|}{\color{blue}30.37/0.6594} & \multicolumn{1}{c|}{\color{blue}33.57/0.7310} & \multicolumn{1}{c}{\color{blue}35.83/0.7772} \\
\multicolumn{1}{l|}{COAST}    & \multicolumn{1}{c|}{}     & \multicolumn{1}{c|}{\color{red} 28.67/0.8623} & \multicolumn{1}{c|}{\color{red}  32.64/0.9261} & \multicolumn{1}{c|}{\color{red}  35.16/0.9508} & \multicolumn{1}{c|}{\color{red}  37.17/0.9645} & \multicolumn{1}{c}{\color{red}  39.00/0.9742 } \\ \shline
\end{tabular}
\end{table*}

\begin{figure}[t]
\setlength{\abovecaptionskip}{0.cm}
\setlength{\belowcaptionskip}{1cm}
\centering
\includegraphics[width=1\linewidth]{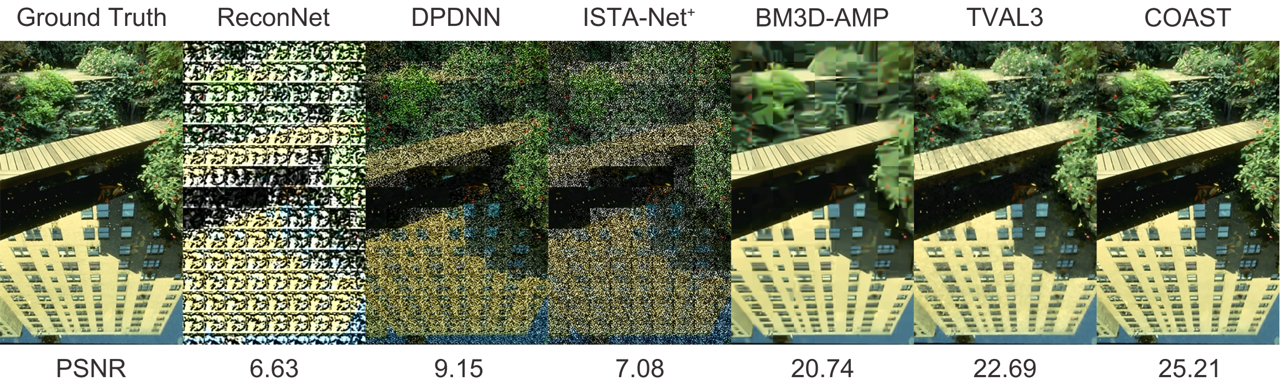}
\caption{Visual comparison on BSD68 dataset of various CS methods on unseen sampling matrix (USM) in the case of $\gamma = 30\% $.}
\label{fig:unseen30}
\end{figure}

\textbf{Comparison on USM:}
To verify the generalization capability of our COAST, we further use the $\mathcal{RPA}$ operator under the condition of FRGM to generate several unseen sampling matrices with $\gamma \in \{10\%,20\%,30\%,40\%,50\%\}$, \textit{i.e.,} $\mathbf{\Phi}^c_{109 \times 1089}$, $\mathbf{\Phi}^c_{218 \times 1089}$, $\mathbf{\Phi}^c_{327 \times 1089 }$, $\mathbf{\Phi}^c_{436 \times 1089}$, $\mathbf{\Phi}^c_{545 \times 1089}$, which do not belong to $\{\mathbf{\Phi}_t\}_{t=1}^{125}$. Without additional training, we directly employ the learned models in Table~\ref{tab:result_noise_0} for these unseen sampling matrices and show the PSNR and SSIM results of Set11 dataset in Table~\ref{tab:diff_seen_unseen} for various CS ratios. 
Undoubtedly, the $\mathbf{\Phi}$-specific ISTA-Net$^+$ \cite{zhang2018ista}, ReconNet \cite{lohit2018convolutional} and DPDNN \cite{DBLP:journals/pami/DongWYSWL19} can not be generalized to arbitrary unseen sampling matrices and perform very poorly, while our COAST outperforms the computationally expensive TVAL3 and BM3D-AMP and still yields impressive performance on par with the sampling matrix in $\{\mathbf{\Phi}_t\}_{t=1}^{125}$, exhibiting good generalization capability. Fig.~\ref{fig:unseen50} and Fig.~\ref{fig:unseen30} provide the visual comparison in the case of USM for several methods on two test images in BSD68 when CS ratio is 30\% and 50\%, respectively, which clearly shows that our COAST is not affected and produces best results. Moreover, Table~\ref{tab:RPA2} reports the average PSNR results for four other unseen sampling matrices with $\gamma \notin \{10\%,20\%,30\%,40\%,50\%\}$, showing similar performance to the model separately trained for each one, which again demonstrates the effectiveness and generality of our COAST to deal with arbitrary sampling matrices.

\begin{table}[t]
\centering
\caption{Average PSNR performance of our COAST on USM with unseen CS ratios ( $\gamma \notin \{10\%,20\%,30\%,40\%,50\%\}$).}
\label{tab:RPA2}
\begin{tabular}{c|c|c|c|c}
\shline 
Matrix & $\mathbf{\Phi}^c_{261 \times 1089}$ &  $\mathbf{\Phi}^c_{294 \times 1089}$     &    $\mathbf{\Phi}^c_{359 \times 1089}$   &  $\mathbf{\Phi}^c_{392 \times 1089}$      \\ \shline 
PSNR & 33.63                               & 35.64 & 36.99 & 37.65 \\ \shline
\end{tabular}
\end{table}

\begin{table}[t]
    \centering
    \footnotesize 
    \setlength{\tabcolsep}{3pt}
    \caption{Reconstruction performance (PSNR) between COAST-A and COAST-S in the cases of various CS ratios and patch sizes on the Set11 dataset.} 
    \label{tab:main_result}
    \begin{tabular}{@{}l c c c c c c c c@{}}
    \shline
    \multicolumn{1}{l|}{} & \multicolumn{3}{c|}{Arbitrary $\gamma$ } & \multicolumn{5}{c}{Arbitrary $\sqrt{N}$}  \\ \hline  
\multicolumn{1}{l|}{CS Ratio $\gamma$} &0.1 & 0.3 & \multicolumn{1}{c|}{0.5} & 0.3 & 0.3 & 0.3 & 0.3 & \multicolumn{1}{c}{0.3} \\ 
   \hline
  \multicolumn{1}{l|}{Patch Size $\sqrt{N}$} & 33  & 33  & \multicolumn{1}{c|}{33}  & 32  & 33  & 36  & 40  & \multicolumn{1}{c}{48} \\
    \shline 
    \multicolumn{1}{l|}{COAST-S}   & 28.67 & 35.14 & \multicolumn{1}{c|}{39.18} & 34.83 & 35.14 &35.30 &35.54  & \multicolumn{1}{c}{35.65} \\

    \multicolumn{1}{l|}{COAST-A} &28.71  & 35.10 & \multicolumn{1}{c|}{38.99} &34.83  & 35.10 &35.27 & 35.49 & \multicolumn{1}{c}{35.60} \\  \hline
    
     \multicolumn{1}{l|}{PSNR Distance} &+0.04  & -0.04 &\multicolumn{1}{c|}{-0.19} &0.00  & -0.04 &-0.03 &-0.05 & \multicolumn{1}{c}{-0.05} \\

    \shline
    \end{tabular}

\end{table}

\begin{table}[t]
    \centering

    \footnotesize 
    \caption{Reconstruction performance (PSNR) between COAST-N and COAST-S in the cases of various CS ratios and noise levels on the Set11 dataset.} 
    \label{tab:noise_result}
    \begin{tabular}{@{}l c c c c c c@{}}
    \shline
    \multicolumn{1}{l|}{} & \multicolumn{6}{c}{Arbitrary $\sigma$} \\
\hline 
\multicolumn{1}{l|}{CS Ratio $\gamma$} & 0.1 & 0.3 & 0.5 & 0.1 & 0.3 & \multicolumn{1}{c}{0.5} \\ 
    \hline \multicolumn{1}{l|}{Noise Level $\sigma$} & 5 & 5 & 5 & 10 & 10 & \multicolumn{1}{c}{10} \\
    \shline 
    \multicolumn{1}{l|}{COAST-S}  & 27.78   & 32.82 & 35.14 & 26.68 & 30.73 & \multicolumn{1}{c}{32.50} \\

    \multicolumn{1}{l|}{COAST-N} & 27.88   & 32.76 & 35.09 & 26.67 & 30.66 & \multicolumn{1}{c}{32.46} \\ \hline 
    
     \multicolumn{1}{l|}{PSNR Distance} & +0.10   &-0.06  & -0.05 & -0.01 & -0.07 & \multicolumn{1}{c}{-0.04}  \\

    \shline
    \end{tabular}

\end{table}

\subsection{Single Model for Arbitrary Sampling Matrices}
\label{single_model}
In order to verify the effectiveness and robustness of our COAST to handle arbitrary sampling matrices with one single model, we further carry out CS reconstruction experiments from noiseless and noisy CS measurements in the cases of various CS ratios, patch sizes, and noise levels for FRGM.

\textbf{Reconstruction from noiseless CS measurements:} Define a set of CS ratios as $\mathbf{\Gamma}=\{10\%,20\%,30\%,40\%,50\%\}$ and a set of patch sizes as $\mathbf{P}=\{32^2, 33^2, 36^2, 40^2, 48^2\}$. Then, we generate a set of FRGMs, denoted by $\mathbf{\Omega}=\{\mathbf{\Phi}_{M_i\times N_i}\}$, where $\frac{M_i}{N_i}\in \mathbf{\Gamma}$ and $N_i\in \mathbf{P}, 1 \leq i \leq 25$. Obviously, $\mathbf{\Omega}$ has 25 different sampling matrices with various dimensions. Here, the COAST model trained by the augmented set of $\mathbf{\Omega}$, \textit{i.e.,} $\mathcal{RPA}(\mathbf{\Omega})$ (including $25\times N_S$ sampling matrices) is referred to as \textbf{COAST-A}, while each COAST model separately trained by the augmented set of each $\mathbf{\Phi}_{M_i\times N_i}$, \textit{i.e.},     $\mathcal{RPA}(\mathbf{\Phi}_{M_i\times N_i})$ (including $N_S$ sampling matrices) is referred to as \textbf{COAST-S}. $N_S$ is set to be $25$ and $\sigma$ is set to be 0.
\begin{figure*}[t]
\setlength{\abovecaptionskip}{0.cm}
\setlength{\belowcaptionskip}{1cm}
\centering
\includegraphics[width=1\linewidth]{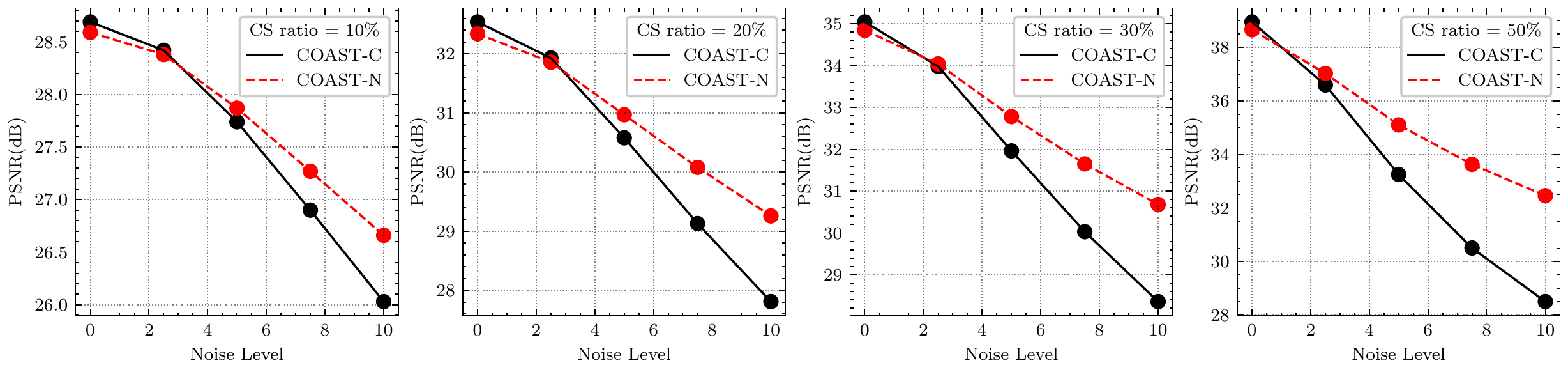}
\caption{Performance comparison on the robustness to the Gaussian noise between COAST-C and COAST-N on Set11 dataset.}
\label{fig: ablation_noise}
\vspace{-0pt}
\end{figure*}

The PSNR results between COAST-A and COAST-S in the cases of various CS ratios and patch sizes on the Set11 dataset are reported in Table~\ref{tab:main_result}. One can clearly observe that COAST-A with a single model is able to handle various arbitrary CS ratios and arbitrary patch sizes effectively, and achieves CS performance on par with COAST-S, with PSNR distances being all below 0.2 dB. Surprisingly, COAST-A can even achieve better performance than the corresponding single model COAST-S in the case of $\gamma = 10\% $ and $\sqrt{N}$ = 33. Table~\ref{tab:main_result} fully verifies the effectiveness and generalization of our COAST to deal with arbitrary sampling matrices.

\begin{table*}[t]
\centering
\caption{Ablation study of different components. The experiments are conducted on FRGM with different CS ratios on the Set11 dataset.} 
\label{tab:ablation_study}
\begin{tabular}{c|c|c|c|c|c|c|c|c|c|c}
\shline
\multicolumn{1}{c|}{\multirow{2}{*}{Setting}} &
\multicolumn{1}{c|}{\multirow{2}{*}{RPA}} &
\multicolumn{1}{c|}{\multirow{2}{*}{CU}}  & \multicolumn{1}{c|}{\multirow{2}{*}{Sharing}} & \multicolumn{1}{c|}{\multirow{2}{*}{PnP-D}} & \multicolumn{1}{c|}{\multirow{2}{*}{Parameters}} & \multicolumn{5}{c}{CS Ratio $\gamma$}  
\\ \cline{7-11} \multicolumn{1}{c|}{}   & \multicolumn{1}{c|}{} & \multicolumn{1}{c|}{} & \multicolumn{1}{c|}{} & \multicolumn{1}{c|}{} & \multicolumn{1}{c|}{}  & 10\% & 20\% & 30\% & 40\% & 50\% \\ \shline
(a)      & \ding{55} & \ding{55}       & \ding{55}       & \ding{55}          & 1,121,960  & 27.04           & 31.25           & 33.91          & 36.15           & 38.18           \\ 
(b)      & $\surd$  & \ding{55}       & \ding{55}       & \ding{55}          & 1,121,960  & 27.69           & 31.82           & 34.45           & 36.58           & 38.43           \\ 
(c)      & $\surd$& $\surd$ & \ding{55}       & \ding{55}          & 1,127,720  & 27.74           & 31.87           & 34.48           & 36.65           & 38.56           \\ 
(d)      & $\surd$ & $\surd$ & $\surd$ & \ding{55}          & 1,122,056  & 27.76           & 31.86           & 34.48           & 36.66           & 38.58           \\ 
(e)      & $\surd$ & $\surd$ & $\surd$ & $\surd$    & 1,122,056  & \textbf{28.69}  & \textbf{32.54}  & \textbf{35.04}  & \textbf{37.13}  & \textbf{38.94}  \\ \shline
\end{tabular}
\end{table*}

\begin{table}[t]
    \centering
    \footnotesize 
    \caption{Average PSNR results of our COAST trained with various values of $N_S$ in the case of $\gamma=30\%$ for FRGM and DALM.} 
    \label{tab:RPA1}
    \begin{tabular}{@{}l c c c c c c@{}}
       \shline
       \multicolumn{1}{l|}{\multirow{2}{*}{Matrix Type}} & \multicolumn{6}{c}{$N_S$}
       \\ \cline{2-7}
       \multicolumn{1}{l|}{} &
       \multicolumn{1}{c|}{1} &
       \multicolumn{1}{c|}{5} &
       \multicolumn{1}{c|}{10} &
       \multicolumn{1}{c|}{15} &
       \multicolumn{1}{c|}{20} &
       \multicolumn{1}{c}{25}

       \\ \shline 
        \multicolumn{1}{c|}{FRGM} &
        \multicolumn{1}{c|}{34.34} &
        \multicolumn{1}{c|}{34.87} &
        \multicolumn{1}{c|}{35.03} &
        \multicolumn{1}{c|}{35.07} &
        \multicolumn{1}{c|}{35.12} &
        \multicolumn{1}{c}{\bf 35.14}
    \\  
        \multicolumn{1}{c|}{DALM} &
        \multicolumn{1}{c|}{35.89} &
        \multicolumn{1}{c|}{\bf 36.18} &
        \multicolumn{1}{c|}{36.17} &
        \multicolumn{1}{c|}{\bf 36.18} &
        \multicolumn{1}{c|}{-} &
        \multicolumn{1}{c}{-}
    \\ \shline
        
    \end{tabular}
    \vspace{0pt}
    \vspace{-6pt}
\end{table}

\textbf{Reconstruction from noisy CS measurements:}
\label{sec:noise}
To further demonstrate the robustness of our COAST to noise, we conduct CS reconstruction experiments from noisy CS measurements, \textit{i.e.,} the noise deviation $\sigma$ of the additive white Gaussian measurement noise $\mathbf{n}$ in Eq.~(\ref{eq: CS}) is not set to be zero. Define a set of CS ratios as $\mathbf{\Gamma}=\{10\%,20\%,30\%,40\%,50\%\}$. At training, we adopt the same five sampling matrices in Table~\ref{tab:result_noise_0}, which constitute a set, dubbed $\mathbf{\Omega}=\{\mathbf{\Phi}_{M_i, N}\}$, where $1\leq i \leq 5, N=1089, \frac{M_i}{N} \in \mathbf{\Gamma}$.
The augmented set of $\mathbf{\Omega}$, \textit{i.e.,} $\mathcal{RPA}(\mathbf{\Omega})$ (including $5\times N_S$ sampling matrices) with $\sigma=0$, are used to train our model, which is referred to as \textbf{COAST-C}. $N_S$ is set to be 25.
In addition, the COAST model trained by $\mathcal{RPA}(\mathbf{\Omega})$ and $\sigma \in [0, 10]$ (\textit{i.e.,} $\sigma$ is randomly uniform-sampled from $[0, 10]$) is referred to as \textbf{COAST-N}.
Similar to the \textbf{COAST-S} in the previous noiseless experiment, we separately employ the augmented set of each $\mathbf{\Phi}_{M_i\times N}$, \textit{i.e.},     $\mathcal{RPA}(\mathbf{\Phi}_{M_i\times N})$ (including $N_S$ sampling matrices) with $\sigma = 5/10$ to train our model and denote it by \textbf{COAST-S}.

The PSNR results between COAST-S and COAST-N in the cases of different noise levels and CS ratios on the Set11 dataset are reported in Table~\ref{tab:noise_result}. As we can see, COAST-N with one single model is quite robust to noise and achieves almost the same performance as the corresponding separately trained model COAST-S, with their PSNR distance being less than 0.1dB for all the tasks. More encouragingly, in the case of $\gamma = 10\% $ and $\sigma$  = 5, COAST-N even surpasses COAST-S. In Fig.~\ref{fig: ablation_noise}, we compare the performance between COAST-N and COAST-C in the cases of various noise levels and CS ratios. One can clearly observe that  COAST-N achieves better results than COAST-C as the noise level increases, which further verifies the robustness and effectiveness of our COAST.

\subsection{Ablation Studies and Discussions}

This subsection will present the ablation study to investigate the contribution of each component in our proposed COAST on the default settings of FRGM.

\textbf{Effect of RPA:} As a core component in COAST, random projection augmentation (RPA) strategy augments the training in the sampling space and enjoys two main advantages: promoting performance and promoting generalization.
The experiments in Subsection~\ref{sota} on unseen sampling matrix (USM) have fully verified the generalization ability of our COAST. Here, we mainly evaluate the advantage of promoting performance.
We first implement COAST to conduct experiments on two types of sampling matrices, \textit{i.e.,} fixed random Gaussian matrix ({FRGM}) and data-driven adaptively learned matrix ({DALM})\footnote{It is also worth noticing that unless otherwise specified, the following experiments are conducted on FRGM.}. Table~\ref{tab:RPA1} reports the average PSNR results of our COAST trained with various values of $N_S$ in the case of $\gamma=30\%$ for FRGM and DALM. Note that $N_S=1$ means no additional sampling matrix is introduced and the RPA strategy is not used. From Table~\ref{tab:RPA1}, one can clearly observe that the increase of $N_S$ greatly promotes the reconstruction performance on both FRGM and DALM. In particular, $N_S=25$ obtains 0.8 dB gains over $N_S=1$ on FRGM, while $N_S=15$ obtains 0.29 dB gains over $N_S=1$ on DALM. Besides, the PSNR value hardly increases when $N_S=25$ on FRGM and $N_S=5$ on DALM, respectively. 
Considering the trade-off between memory-saving and recovery performance, we set the default value of $N_S$ to be 25/5 for FRGM/DALM. Moreover, comparing Settings (a) and (b) in Table~\ref{tab:ablation_study}, we observe that RPA greatly promotes CS performance across all the cases.

\textbf{Effect of CU in CPMM:}
The controllable proximal mapping module (CPMM) is able to dynamically modulate the network features and greatly promote the network robustness by adopting the controllable unit (CU). The experiments in  Subsection~\ref{single_model} on reconstruction from noisy CS measurements have substantially proved that CU has the advantage of high noise robustness. Here, we mainly emphasize that the utilization of CU can also bring about performance improvement. Settings (b) and (c) in Table~\ref{tab:ablation_study} provide the performance comparison between CPMM with CU and CPMM without CU for five CS ratios on Set11. One can observe that CPMM with CU consistently outperforms CPMM without CU across all the ratios and obtains on average 0.07 dB gains.
\begin{figure}[t]
\setlength{\abovecaptionskip}{0.cm}
\setlength{\belowcaptionskip}{1cm}
\centering
\includegraphics[width=0.95\linewidth]{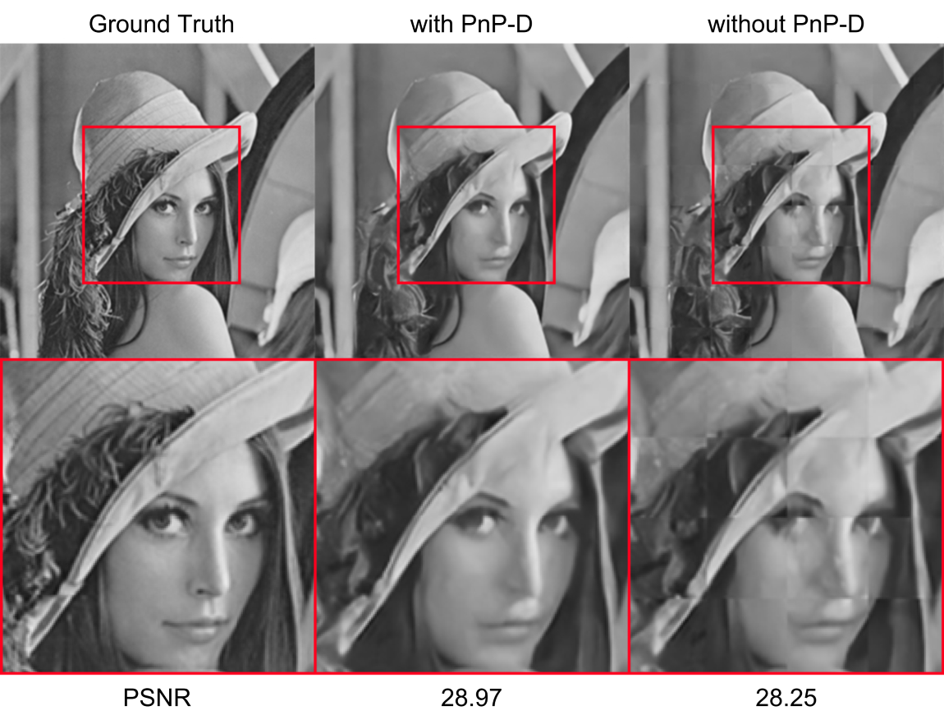} 
\caption{Visual comparison between w/ PnP-D and w/o PnP-D. Obviously, w/ PnP-D achieves better results than w/o PnP-D by effectively eliminating the blocking artifacts.}
\vspace{-5pt}
\label{fig:ablation_study}
\end{figure}

\textbf{Shared vs. Unshared:}
In our experiment, the recovery subnet (RS) of our COAST has 20 phases and each phase has 3 CUs. Thus, there exist 60 CUs in total in COAST. Here, we study the effect of the parameter sharing across all the CUs. Note that each CU has $2\times 32 + 32 = 96$ parameters. Therefore, the shared strategy successfully reduces the parameter number from $96\times 60=5760$ to $96$. Settings (c) and (d) in Table~\ref{tab:ablation_study} provide the PSNR comparison between unshared strategy and shared strategy. Obviously, shared strategy achieves almost the same performance as unshared strategy, which fully demonstrates the effectiveness of our shared strategy, greatly reducing the parameter number without performance loss.

\textbf{With vs. Without PnP-D:}
The plug-and-play deblocking (PnP-D) strategy can be easily integrated into existing CS frameworks. Settings (d) and (e) in Table~\ref{tab:ablation_study} give the PSNR comparison between w/ PnP-D and w/o PnP-D. It is clear to see that PnP-D strategy greatly boosts the performance across all ratios, with the most significant improvement up to 0.93 dB, which fully verifies the effectiveness of PnP-D. Fig.~\ref{fig:ablation_study} further shows the visual results by w/ PnP-D and w/o PnP-D, which clearly demonstrates that the PnP-D is able to eliminate the blocking artifact effectively, resulting in high-quality reconstructed images.

\begin{figure}[t]
\setlength{\abovecaptionskip}{0.cm}
\setlength{\belowcaptionskip}{1cm}
\centering
\includegraphics[width=0.95\linewidth]{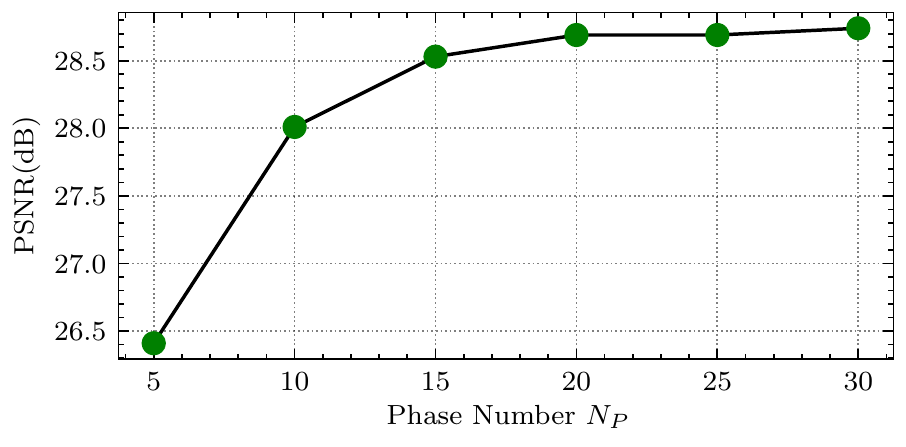} 
\caption{Average PSNR curves on Set11 achieved by COAST with various phase numbers in the case of $\gamma = 10\% $.}
\vspace{-8pt}
\label{fig:phase}
\end{figure}

\textbf{Performance of Phase Number:}
Since each phase of COAST corresponds to one iteration in traditional ISTA, it is expected that larger phase number $N_P$ will lead to higher reconstruction performance. Fig.~\ref{fig:phase} shows the average PSNR curve by COAST for Set11 with respect to different phase numbers in the case of $\gamma=10\% $. One can observe that the PSNR curve increases as $N_P$ increases. However, the curve is becoming almost flat when $N_P\geq20$. Thus, considering the trade-off between computational complexity and reconstruction performance, we set the default value of phase number $N_P$ to be 20 in COAST.  

\textbf{Discussion:}
As far as we know, there exist three methods with the ability of handling arbitrary CS ratios, \textit{i.e.}, SCSNet \cite{shi2019scalable}, RACS \cite{lohit2018rate} and DRNN \cite{xu2020compressed}. We compare them with our COAST in the following paragraphs.

Although SCSNet \cite{shi2019scalable} and RACS \cite{lohit2018rate} can also deal with arbitrary sampling ratios using only one model, there exist two main differences between their methods and our COAST: scope of application w.r.t sampling matrices and mechanism of handling arbitrary CS ratios. Firstly, our COAST is generally applicable whenever a sampling matrix is given \textit{beforehand} (including FRGM and DALM), while SCSNet and RACS \textit{learn} the sampling matrix and are not applicable when one wants to use any given sampling matrix, whether it is learned or not. Secondly, the mechanisms of how to enable the network to handle arbitrary CS ratios are different. Specifically, SCSNet achieves sampling and reconstruction at any sampling ratio by using a greedy method to select the measurement bases. RACS trains a single-ratio reconstruction network first, and then `adapts' new sampling matrices to the parameter-fixed reconstruction network. For our COAST, the proposed RPA enables our reconstruction network to jointly learn different knowledge from multiple sampling matrices, thus enabling our COAST to robustly deal with arbitrary CS ratios.

As for DRNN \cite{xu2020compressed}, two main aspects differentiate DRNN and our COAST: application field and object of operation. Firstly, 
DRNN aims to perform \textit{classification} directly on compressive measurements, while our COAST aims to \textit{recover} the unknown natural signal from its compressed measurements. Secondly, the training scheme of DRNN can be considered as a \textit{ratio-level} operation, and ignores the \textit{matrix-level}. Different from DRNN, our proposed RPA is not limited to \textit{ratio-level}, but further promotes the performance of each sampling matrix by adopting the \textit{matrix-level} argumentation, which is clearly verified in Table~\ref{tab:RPA1} in our paper.

\begin{table*}[]
\caption{Average PSNR comparisons for ISTA-Net$^{+}$(zero initialized) \cite{zhang2018ista}, BM3D-AMP \cite{metzler2016denoising} and our COAST on \textbf{unseen} sampling matrix (USM) across different CS ratios on Set11 dataset. The number in the grey box indicates the PSNR gain or drop compared with corresponding \textbf{seen} sampling matrices. For ISTA-Net$^{+}$, unseen sampling matrices are further classified into 2 categories: the unseen ratios (\underline{underlined}) and the seen ratio.}
\setlength{\tabcolsep}{5pt}  
\label{tab:usm_phitb}
\footnotesize
\centering
\begin{tabular}{c|c|c|c|c|c|c}
\shline
\multirow{2}{*}{Methods}    & \multirow{2}{*}{\begin{tabular}[c]{@{}c@{}}Train CS Ratio\\ (range)\end{tabular}} & \multicolumn{5}{c}{Test on Unseen Sampling Matrix (USM) }                     \\ \cline{3-7}
                            &                                                                                      & 10\%         & 20\%         & 30\%         & 40\%         & 50\%                                                                                                \\ \shline
\multirow{5}{*}{ISTA-Net$^{+}$} & 10\%                                                                                 & 20.67\colorbox{gray!20}{($\textcolor{black}{\downarrow}$ 5.56)}  & \underline{20.56}\colorbox{gray!20}{($\textcolor{black}{\downarrow}$ \underline{10.03})}  & \underline{14.31}\colorbox{gray!20}{($\textcolor{black}{\downarrow}$ \underline{19.12})}   & \underline{11.14}\colorbox{gray!20}{($\textcolor{black}{\downarrow}$ \underline{24.63})}   & \underline{10.32}\colorbox{gray!20}{($\textcolor{black}{\downarrow}$ \underline{27.59})}                                                                   \\
                            & 20\%                                                                                 & \underline{16.08}\colorbox{gray!20}{($\textcolor{black}{\downarrow}$ \underline{10.15})}   & 29.04\colorbox{gray!20}{($\textcolor{black}{\downarrow}$ 1.55)}   & \underline{26.43}\colorbox{gray!20}{($\textcolor{black}{\downarrow}$ \underline{7.00})}     & \underline{19.07}\colorbox{gray!20}{($\textcolor{black}{\downarrow}$ \underline{16.70})}     & \underline{13.25}\colorbox{gray!20}{($\textcolor{black}{\downarrow}$ \underline{24.66})}                                                                              \\
                            & 30\%                                                                                 & \underline{12.39}\colorbox{gray!20}{($\textcolor{black}{\downarrow}$ \underline{13.84})}      & \underline{22.92}\colorbox{gray!20}{($\textcolor{black}{\downarrow}$ \underline{7.67})}       & 32.65\colorbox{gray!20}{($\textcolor{black}{\downarrow}$ 0.78)}     & \underline{34.63}\colorbox{gray!20}{($\textcolor{black}{\downarrow}$ \underline{1.14})}      & \underline{35.63}\colorbox{gray!20}{($\textcolor{black}{\downarrow}$ \underline{2.28})}                                                                                \\
                            & 40\%                                                                                 & \underline{10.95}\colorbox{gray!20}{($\textcolor{black}{\downarrow}$ \underline{15.28})}        & \underline{18.37}\colorbox{gray!20}{($\textcolor{black}{\downarrow}$ \underline{12.22})}       & \underline{30.21}\colorbox{gray!20}{($\textcolor{black}{\downarrow}$ \underline{3.22})}       & 35.28\colorbox{gray!20}{($\textcolor{black}{\downarrow}$ 0.49)}     & \underline{37.36}\colorbox{gray!20}{($\textcolor{black}{\downarrow}$ \underline{0.55})}                                                                                   \\
                            & 50\%                                                                                 & \underline{9.38}\colorbox{gray!20}{($\textcolor{black}{\downarrow}$ \underline{16.85})}      & \underline{13.82}\colorbox{gray!20}{($\textcolor{black}{\downarrow}$ \underline{16.77})}  & \underline{20.17}\colorbox{gray!20}{($\textcolor{black}{\downarrow}$ \underline{13.26})}  & \underline{32.98}\colorbox{gray!20}{($\textcolor{black}{\downarrow}$ \underline{2.79})}  & 37.48\colorbox{gray!20}{($\textcolor{black}{\downarrow}$ 0.43)}                                                                           \\ \hline
BM3D-AMP                      & --                                                                              & 22.47\colorbox{gray!20}{($\textcolor{black}{\downarrow}$ 0.13)} & 26.80\colorbox{gray!20}{($\textcolor{black}{\uparrow}$ 0.03)} & 30.37\colorbox{gray!20}{($\textcolor{black}{\uparrow}$ 0.11)} & 33.57\colorbox{gray!20}{($\textcolor{black}{\downarrow}$ 0.09)} & 35.83\colorbox{gray!20}{($\textcolor{black}{\downarrow}$ 0.10)} \\ \hline
COAST                      & 10\%-50\%                                                                              & 28.67\colorbox{gray!20}{($\textcolor{black}{\downarrow}$ 0.02)} & 32.64\colorbox{gray!20}{($\textcolor{black}{\uparrow}$ 0.10)} & 35.16\colorbox{gray!20}{($\textcolor{black}{\uparrow}$ 0.12)} & 37.17\colorbox{gray!20}{($\textcolor{black}{\uparrow}$ 0.04)} & 39.00\colorbox{gray!20}{($\textcolor{black}{\uparrow}$ 0.06)}                                                                                 \\ \shline
\end{tabular}
\end{table*}
\vspace{-7pt}
\subsection{Analysis on USM for Deep Unfolding Methods}
\vspace{-2pt}
To analyze why the unfolding based methods (\textit{e.g.}, ISTA-Net$^+$) fail with unseen sampling matrices (USM), we discuss the dependence of the deep unfolding networks on the specific sampling matrix in the \textit{initialization} process and the \textit{reconstruction} process, respectively.

\textbf{Initialization:} 
To deal with the dimensionality mismatch between the original image and its CS measurements, deep network-based methods require an initialization. Some methods \cite{zhang2018ista} depend on the specific sampling matrix used in training, while some methods \cite{DBLP:conf/nips/MetzlerMB17} don't (\textit{e.g.}, directly initialized by 0). ISTA-Net$^+$, which belongs to the former category, first uses the training dataset and the specific training sampling matrix to compute a linear mapping matrix $\mathbf{Q}_{init}$, and then obtains $\mathbf{\hat{x}}^{(0)}$=$\mathbf{Q}_{init}\mathbf{y}$.
Due to the fixed dimension of $\mathbf{Q}_{init}$ in initialization, ISTA-Net$^+$ can not handle sampling matrices in \textit{unseen} CS ratios. When handling unseen sampling matrices in \textit{seen} CS ratio, ISTA-Net$^+$ still performs very poorly, as shown in Table~\ref{tab:diff_seen_unseen}.

Based on the above analysis, one may wonder whether deep unfolding methods would still fail with USM when removing the dependence on the sampling matrix in initialization process.

\textbf{Reconstruction:} To eliminate the effect of initialization and focus on how the reconstruction process influences the generalization ability on USM, we initially set zero values in ISTA-Net$^{+}$ and compare the average PSNR results for ISTA-Net$^{+}$ \cite{zhang2018ista}, BM3D-AMP \cite{metzler2016denoising} and COAST on \textit{unseen} sampling matrices across different CS ratios on Set11 dataset (shown in Table~\ref{tab:usm_phitb}). The number in the grey box indicates the PSNR gain (${\uparrow}$) or drop (${\downarrow}$) compared with corresponding \textit{seen} sampling matrices, which represents how general the method is when being extended to unseen sampling matrices.
We can observe that, for the traditional iterative method BM3D-AMP, the PSNR performance fluctuation shown in grey boxes is mild, indicating a nice generality w.r.t unseen sampling matrices.
However, when applying the deep unfolding network ISTA-Net$^{+}$ to unseen sampling matrices, the PSNR performance drops a lot (see the grey boxes for ISTA-Net$^{+}$), especially on unseen ratios (underlined in Table~\ref{tab:usm_phitb},), which demonstrates that ISTA-Net$^{+}$ exhibits inferior generality of being extended to unseen sampling matrices.
As a comparison, the PSNR performance gain or drop is quite slight for our COAST, thus verifying its superior generality of handling USM.

One might think that, ISTA-Net$^{+}$ is a special case of traditional ISTA when testing, and should be robust to arbitrary sampling matrices just like traditional iterative methods. 
But this is in fact not the case, as compared earlier.
Considering the differences between traditional iterative methods and deep unfolding models, we attribute such failure to two aspects. 

Firstly, the selected image prior (\textit{e.g.}, optimal transform) and the optimization parameters (\textit{e.g.}, step size and regularization parameter) are usually hand-crafted in traditional iterative methods without considering any \textit{specific} sampling matrix, while in deep unfolding models, all the parameters (\textit{e.g.}, optimal transforms, step sizes, shrinkage thresholds, etc.) are learned from training data pairs, which contain the information of a \textit{specific} training sampling matrix. Therefore, these learned parameters are somewhat dependent on the training sampling matrix, and thus leading to inferior generality when being applied to unseen sampling matrices. 

Secondly, although traditional optimization-based methods and deep unfolding models are both iterative, the iteration number is much smaller for the latter. Taking ISTA-Net$^{+}$ as an example, it is forced to converge after 9 iterations, thus leading to relatively larger step sizes and stronger proximal mapping, which indicates fewer and larger iteration steps towards the optimal solution when optimizing the target problem. 
However, when being applied to unseen sampling matrices, the target problem may change a lot, and therefore, larger iteration steps might relatively lead to a suboptimal result, thus resulting in inferior generality w.r.t. unseen sampling matrices.

To sum up, for deep unfolding models, the above two reasons cause the network parameters to `overfit' the specific sampling matrix used in training, and thus leading to inferior generality when handling USM.

\section{Conclusion}

In this paper, a novel \textbf{CO}ntrollable \textbf{A}rbitrary-\textbf{S}ampling ne\textbf{T}work, dubbed \textbf{COAST}, is proposed for compressive sensing under the optimization-inspired deep unfolding framework. COAST is able to deal with arbitrary sampling matrices with one single model, exhibiting good interpretability, high robustness as well as attractive computational speed. Extensive experiments demonstrate that COAST not only achieves CS performance on par with single models separately trained for different specific sampling matrices, but also greatly improves current state-of-the-art CS results on both fixed random Gaussian matrix (FRGM) and data-driven adaptively learned matrix (DALM). It is expected that the proposed COAST will inspire new insights for a wide range of potential real-world applications in computational imaging in the future, including but not limited to handling arbitrary sampling masks in accelerating MRI, dealing with multiple digital micromirror devices (DMDs) in single-pixel imaging and reconstructing under different system matrices in sparse-view CT.
Future work includes the direct extensions of COAST on MRI data and video data applications.

{
\bibliographystyle{IEEEtran}
\bibliography{egbib}

\begin{thebibliography}{10}
\providecommand{\url}[1]{#1}
\csname url@samestyle\endcsname
\providecommand{\newblock}{\relax}
\providecommand{\bibinfo}[2]{#2}
\providecommand{\BIBentrySTDinterwordspacing}{\spaceskip=0pt\relax}
\providecommand{\BIBentryALTinterwordstretchfactor}{4}
\providecommand{\BIBentryALTinterwordspacing}{\spaceskip=\fontdimen2\font plus
\BIBentryALTinterwordstretchfactor\fontdimen3\font minus
  \fontdimen4\font\relax}
\providecommand{\BIBforeignlanguage}[2]{{%
\expandafter\ifx\csname l@#1\endcsname\relax
\typeout{** WARNING: IEEEtran.bst: No hyphenation pattern has been}%
\typeout{** loaded for the language `#1'. Using the pattern for}%
\typeout{** the default language instead.}%
\else
\language=\csname l@#1\endcsname
\fi
#2}}
\providecommand{\BIBdecl}{\relax}
\BIBdecl

\bibitem{duarte2008single}
M.~F. Duarte, M.~A. Davenport, D.~Takbar, J.~N. Laska, T.~Sun, K.~F. Kelly, and
  R.~G. Baraniuk, ``Single-pixel imaging via compressive sampling,'' \emph{IEEE
  Signal Processing Magazine}, vol.~25, no.~2, pp. 83--91, 2008.

\bibitem{rousset2017adaptive}
F.~Rousset, N.~Ducros, A.~Farina, G.~Valentini, C.~D’Andrea, and F.~Peyrin,
  ``Adaptive basis scan by wavelet prediction for single-pixel imaging,''
  \emph{IEEE Transactions on Computational Imaging}, vol.~3, no.~1, pp. 36--46,
  2017.

\bibitem{lustig2007sparse}
M.~Lustig, D.~Donoho, and J.~M. Pauly, ``{Sparse MRI}: The application of
  compressed sensing for rapid mr imaging,'' \emph{Magnetic Resonance in
  Medicine}, vol.~58, no.~6, pp. 1182--1195, 2007.

\bibitem{szczykutowicz2010dual}
T.~P. Szczykutowicz and G.-H. Chen, ``Dual energy ct using slow kvp switching
  acquisition and prior image constrained compressed sensing,'' \emph{Physics
  in Medicine \& Biology}, vol.~55, no.~21, pp. 6411--6429, 2010.

\bibitem{zhang2013compressed}
Z.~Zhang, T.-P. Jung, S.~Makeig, and B.~D. Rao, ``Compressed sensing for
  energy-efficient wireless telemonitoring of noninvasive fetal {ECG} via block
  sparse bayesian learning,'' \emph{IEEE Transactions on Biomedical
  Engineering}, vol.~60, no.~2, pp. 300--309, 2013.

\bibitem{sharma2016application}
S.~K. Sharma, E.~Lagunas, S.~Chatzinotas, and B.~Ottersten, ``Application of
  compressive sensing in cognitive radio communications: A survey,'' \emph{IEEE
  Communications Surveys \& Tutorials}, vol.~18, no.~3, pp. 1838--1860, 2016.

\bibitem{yang2016deep}
Y.~Yang, J.~Sun, H.~Li, and Z.~Xu, ``Deep {ADMM}-net for compressive sensing
  mri,'' in \emph{Proceedings of Advances in Neural Information Processing
  Systems (NIPS)}, 2016, pp. 10--18.

\bibitem{zhao2016video}
C.~Zhao, S.~Ma, J.~Zhang, R.~Xiong, and W.~Gao, ``Video compressive sensing
  reconstruction via reweighted residual sparsity,'' \emph{IEEE Transactions on
  Circuits and Systems for Video Technology}, vol.~27, no.~6, pp. 1182--1195,
  2016.

\bibitem{zhang2018ista}
J.~Zhang and B.~Ghanem, ``{ISTA-Net}: Interpretable optimization-inspired deep
  network for image compressive sensing,'' in \emph{Proceedings of IEEE
  Conference on Computer Vision and Pattern Recognition (CVPR)}, 2018, pp.
  1828--1837.

\bibitem{duarte2009learning}
J.~M. Duarte-Carvajalino and G.~Sapiro, ``Learning to sense sparse signals:
  Simultaneous sensing matrix and sparsifying dictionary optimization,''
  \emph{IEEE Transactions on Image Processing}, vol.~18, no.~7, pp. 1395--1408,
  2009.

\bibitem{hong2018online}
T.~Hong and Z.~Zhu, ``Online learning sensing matrix and sparsifying dictionary
  simultaneously for compressive sensing,'' \emph{Signal Processing}, vol. 153,
  pp. 188--196, 2018.

\bibitem{adler2016deep}
A.~Adler, D.~Boublil, and M.~Zibulevsky, ``Block-based compressed sensing of
  images via deep learning,'' in \emph{Proceedings of International Workshop on
  Multimedia Signal Processing (MMSP)}, 2017, pp. 1--6.

\bibitem{du2018fully}
J.~Du, X.~Xie, C.~Wang, G.~Shi, X.~Xu, and Y.~Wang, ``Fully convolutional
  measurement network for compressive sensing image reconstruction,''
  \emph{Neurocomputing}, vol. 328, pp. 105--112, 2019.

\bibitem{shi2017deep}
W.~Shi, F.~Jiang, S.~Zhang, and D.~Zhao, ``Deep networks for compressed image
  sensing,'' in \emph{Proceedings of IEEE International Conference on
  Multimedia and Expo (ICME)}, 2017, pp. 877--882.

\bibitem{shi2019image}
W.~Shi, F.~Jiang, S.~Liu, and D.~Zhao, ``Image compressed sensing using
  convolutional neural network,'' \emph{IEEE Transactions on Image Processing},
  vol.~29, pp. 375--388, 2019.

\bibitem{lohit2018convolutional}
S.~Lohit, K.~Kulkarni, R.~Kerviche, P.~Turaga, and A.~Ashok, ``Convolutional
  neural networks for noniterative reconstruction of compressively sensed
  images,'' \emph{IEEE Transactions on Computational Imaging}, vol.~4, no.~3,
  pp. 326--340, 2018.

\bibitem{shi2019scalable}
W.~Shi, F.~Jiang, S.~Liu, and D.~Zhao, ``Scalable convolutional neural network
  for image compressed sensing,'' in \emph{Proceedings of the IEEE Conference
  on Computer Vision and Pattern Recognition}, 2019, pp. 12\,290--12\,299.

\bibitem{9199540}
Y.~Sun, J.~Chen, Q.~Liu, B.~Liu, and G.~Guo, ``Dual-path attention network for
  compressed sensing image reconstruction,'' \emph{IEEE Transactions on Image
  Processing}, vol.~29, pp. 9482--9495, 2020.

\bibitem{chen2020learning}
J.~Chen, Y.~Sun, Q.~Liu, and R.~Huang, ``Learning memory augmented cascading
  network for compressed sensing of images.'' in \emph{Proceedings of European
  Conference on Computer Vision (ECCV)}, 2020, pp. 513--529.

\bibitem{gregor2010learning}
K.~Gregor and Y.~LeCun, ``Learning fast approximations of sparse coding,'' in
  \emph{Proceedings of International Conference on Machine Learning (ICML)},
  2010, pp. 399--406.

\bibitem{8753511}
D.~Ren, W.~Zuo, D.~Zhang, L.~Zhang, and M.-H. Yang, ``Simultaneous fidelity and
  regularization learning for image restoration,'' \emph{IEEE Transactions on
  Pattern Analysis and Machine Intelligence}, vol.~43, no.~1, pp. 284--299,
  2021.

\bibitem{DBLP:conf/nips/MetzlerMB17}
C.~A. Metzler, A.~Mousavi, and R.~G. Baraniuk, ``Learned {D-AMP:} principled
  neural network based compressive image recovery,'' in \emph{Proceedings of
  Annual Conference on Neural Information Processing Systems (NeurIPS)}, 2017,
  pp. 1772--1783.

\bibitem{borgerding2017amp}
M.~Borgerding, P.~Schniter, and S.~Rangan, ``{AMP}-inspired deep networks for
  sparse linear inverse problems,'' \emph{IEEE Transactions on Signal
  Processing}, vol.~65, no.~16, pp. 4293--4308, 2017.

\bibitem{dong2019denoising}
W.~Dong, P.~Wang, W.~Yin, G.~Shi, F.~Wu, and X.~Lu, ``Denoising prior driven
  deep neural network for image restoration,'' \emph{IEEE Transactions on
  Pattern Analysis and Machine Intelligence}, vol.~41, no.~10, pp. 2305--2318,
  2019.

\bibitem{blumensath2009iterative}
T.~Blumensath and M.~E. Davies, ``Iterative hard thresholding for compressed
  sensing,'' \emph{Applied and Computational Harmonic Analysis}, vol.~27,
  no.~3, pp. 265--274, 2009.

\bibitem{kim2010compressed}
Y.~Kim, M.~S. Nadar, and A.~Bilgin, ``Compressed sensing using a gaussian scale
  mixtures model in wavelet domain,'' in \emph{Proceedings of IEEE
  International Conference on Image Processing (ICIP)}, 2010, pp. 3365--3368.

\bibitem{he2009exploiting}
L.~He and L.~Carin, ``Exploiting structure in wavelet-based bayesian
  compressive sensing,'' \emph{IEEE Transactions on Signal Processing},
  vol.~57, no.~9, pp. 3488--3497, 2009.

\bibitem{zhang2014imageSP}
J.~Zhang, C.~Zhao, D.~Zhao, and W.~Gao, ``Image compressive sensing recovery
  using adaptively learned sparsifying basis via l0 minimization,''
  \emph{Signal Processing}, vol. 103, pp. 114--126, 2014.

\bibitem{zhang2013improved}
J.~Zhang, S.~Liu, R.~Xiong, S.~Ma, and D.~Zhao, ``Improved total variation
  based image compressive sensing recovery by nonlocal regularization,'' in
  \emph{IEEE International Symposium on Circuits and Systems (ISCAS)}, 2013,
  pp. 2836--2839.

\bibitem{DBLP:journals/tip/ZhangZG14}
J.~Zhang, D.~Zhao, and W.~Gao, ``Group-based sparse representation for image
  restoration,'' \emph{{IEEE} Trans. Image Process.}, vol.~23, no.~8, pp.
  3336--3351, 2014.

\bibitem{DBLP:journals/esticas/ZhangZZXMG12}
J.~Zhang, D.~Zhao, C.~Zhao, R.~Xiong, S.~Ma, and W.~Gao, ``Image compressive
  sensing recovery via collaborative sparsity,'' \emph{{IEEE} J. Emerg. Sel.
  Topics Circuits Syst.}, vol.~2, no.~3, pp. 380--391, 2012.

\bibitem{DBLP:conf/dcc/ZhaoZM016}
C.~Zhao, J.~Zhang, S.~Ma, and W.~Gao, ``Nonconvex {Lp} nuclear norm based
  {ADMM} framework for compressed sensing,'' in \emph{Proceedings of Data
  Compression Conference (DCC)}, 2016, pp. 161--170.

\bibitem{dong2014compressive}
W.~Dong, G.~Shi, X.~Li, Y.~Ma, and F.~Huang, ``Compressive sensing via nonlocal
  low-rank regularization,'' \emph{IEEE Transactions on Image Processing},
  vol.~23, no.~8, pp. 3618--3632, 2014.

\bibitem{zhang2017learning}
K.~Zhang, W.~Zuo, S.~Gu, and L.~Zhang, ``Learning deep {CNN} denoiser prior for
  image restoration,'' \emph{Proceedings of IEEE Conference on Computer Vision
  and Pattern Recognition (CVPR)}, 2017.

\bibitem{chang2017projector}
J.~H.~R. Chang, C.-L. Li, B.~P{\'o}czos, B.~V. K.~V. Kumar, and A.~C.
  Sankaranarayanan, ``One network to solve them all --- solving linear inverse
  problems using deep projection models,'' \emph{Proceedings of IEEE
  International Conference on Computer Vision (ICCV)}, 2017.

\bibitem{zhao2018cream}
C.~Zhao, J.~Zhang, R.~Wang, and W.~Gao, ``{CREAM}:{CNN-REgularized ADMM}
  framework for compressive-sensed image reconstruction,'' \emph{IEEE Access},
  vol.~6, pp. 76\,838--76\,853, 2018.

\bibitem{li2013efficient}
C.~Li, W.~Yin, H.~Jiang, and Y.~Zhang, ``An efficient augmented lagrangian
  method with applications to total variation minimization,''
  \emph{Computational Optimization and Applications}, vol.~56, no.~3, pp.
  507--530, 2013.

\bibitem{metzler2016denoising}
C.~A. Metzler, A.~Maleki, and R.~G. Baraniuk, ``From denoising to compressed
  sensing,'' \emph{IEEE Transactions on Information Theory}, vol.~62, no.~9,
  pp. 5117--5144, 2016.

\bibitem{DBLP:conf/allerton/MousaviPB15}
A.~Mousavi, A.~B. Patel, and R.~G. Baraniuk, ``A deep learning approach to
  structured signal recovery,'' in \emph{Proceedings of Annual Allerton
  Conference on Communication, Control, and Computing (Allerton)}, 2015, pp.
  1336--1343.

\bibitem{DBLP:journals/dsp/IliadisSK18}
M.~Iliadis, L.~Spinoulas, and A.~K. Katsaggelos, ``Deep fully-connected
  networks for video compressive sensing,'' \emph{Digit. Signal Process.},
  vol.~72, pp. 9--18, 2018.

\bibitem{kulkarni2016reconnet}
K.~Kulkarni, S.~Lohit, P.~Turaga, R.~Kerviche, and A.~Ashok, ``Recon{N}et:
  Non-iterative reconstruction of images from compressively sensed
  measurements,'' in \emph{Proceedings of IEEE Conference on Computer Vision
  and Pattern Recognition (CVPR)}, 2016, pp. 449--458.

\bibitem{DBLP:journals/pami/DongWYSWL19}
W.~Dong, P.~Wang, W.~Yin, G.~Shi, F.~Wu, and X.~Lu, ``Denoising prior driven
  deep neural network for image restoration,'' \emph{{IEEE} Trans. Pattern
  Anal. Mach. Intell.}, vol.~41, no.~10, pp. 2305--2318, 2019.

\bibitem{9019857}
J.~{Zhang}, C.~{Zhao}, and W.~{Gao}, ``Optimization-inspired compact deep
  compressive sensing,'' \emph{IEEE Journal of Selected Topics in Signal
  Processing}, vol.~14, no.~4, pp. 765--774, 2020.

\bibitem{ulyanov2018deep}
D.~Ulyanov, A.~Vedaldi, and V.~Lempitsky, ``Deep image prior,'' in
  \emph{Proceedings of IEEE Conference on Computer Vision and Pattern
  Recognition}, 2018, pp. 9446--9454.

\bibitem{8999514}
Y.~Sun, Y.~Yang, Q.~Liu, J.~Chen, X.-T. Yuan, and G.~Guo, ``Learning
  non-locally regularized compressed sensing network with half-quadratic
  splitting,'' \emph{IEEE Transactions on Multimedia}, vol.~22, no.~12, pp.
  3236--3248, 2020.

\bibitem{DBLP:conf/iccv/YunHCOYC19}
S.~Yun, D.~Han, S.~Chun, S.~J. Oh, Y.~Yoo, and J.~Choe, ``Cutmix:
  Regularization strategy to train strong classifiers with localizable
  features,'' in \emph{Proceedings of IEEE International Conference on Computer
  Vision (ICCV)}, 2019, pp. 6022--6031.

\bibitem{gilton2019neumann}
D.~Gilton, G.~Ongie, and R.~Willett, ``Neumann networks for linear inverse
  problems in imaging,'' \emph{IEEE Transactions on Computational Imaging},
  vol.~6, pp. 328--343, 2019.

\bibitem{DBLP:conf/eccv/DongLHT14}
C.~Dong, C.~C. Loy, K.~He, and X.~Tang, ``Learning a deep convolutional network
  for image super-resolution,'' in \emph{Proceedings of European Conference on
  Computer Vision (ECCV)}, 2014, pp. 184--199.

\bibitem{pytorch}
A.~Paszke, S.~Gross, F.~Massa \emph{et~al.}, ``Pytorch: An imperative style,
  high-performance deep learning library,'' in \emph{Proceedings of Annual
  Conference on Neural Information Processing Systems (NeurIPS)}, 2019, pp.
  8024--8035.

\bibitem{kingma2015adam}
D.~P. Kingma and J.~Ba, ``Adam: A method for stochastic optimization,'' in
  \emph{Proceedings of International Conference on Learning Representations
  (ICLR)}, 2015, pp. 1--15.

\bibitem{martin2001database}
D.~Martin, C.~Fowlkes, D.~Tal, and J.~Malik, ``A database of human segmented
  natural images and its application to evaluating segmentation algorithms and
  measuring ecological statistics,'' in \emph{Proceedings of IEEE International
  Conference on Computer Vision (ICCV)}, 2001, pp. 416--423.

\bibitem{wang2004image}
Z.~Wang, A.~C. Bovik, H.~R. Sheikh, and E.~P. Simoncelli, ``Image quality
  assessment: from error visibility to structural similarity,'' \emph{IEEE
  Transactions on Image Processing}, vol.~13, no.~4, pp. 600--612, 2004.

\bibitem{lohit2018rate}
S.~Lohit, R.~Singh, K.~Kulkarni, and P.~Turaga, ``Rate-adaptive neural networks
  for spatial multiplexers,'' \emph{arXiv preprint arXiv:1809.02850}, 2018.

\bibitem{xu2020compressed}
Y.~Xu, W.~Liu, and K.~F. Kelly, ``Compressed domain image classification using
  a dynamic-rate neural network,'' \emph{IEEE Access}, vol.~8, pp.
  217\,711--217\,722, 2020.

\end{thebibliography}
}

%




\end{document}